\documentclass[10pt,twocolumn,letterpaper]{article}

\usepackage{cvpr}              

\usepackage[utf8]{inputenc} 
\usepackage[T1]{fontenc}    
\usepackage{url}            
\usepackage{booktabs}       
\usepackage{amsfonts}       
\usepackage{nicefrac}       
\usepackage{microtype}      
\usepackage{xcolor}         
\usepackage[accsupp]{axessibility}

\usepackage{multirow}
\usepackage{multicol}
\usepackage{colortbl}
\definecolor{dark-gray}{gray}{0.30}
\newcommand{\pub}[1]{{\color{dark-gray}{\scriptsize{[{#1}]}}}}
\usepackage{graphicx}
\usepackage{subcaption}
\usepackage{array}

\definecolor{cvprblue}{rgb}{0.21,0.49,0.74}
\usepackage[pagebackref,breaklinks,colorlinks,allcolors=cvprblue]{hyperref}
\def\paperID{*****} 
\def\confName{CVPR}
\def\confYear{2026}

\title{Learning Vision-Language-Action World Models for Autonomous Driving}

\author{
    Guoqing Wang$^{1}$ \quad  Pin Tang$^{1}$  \quad Xiangxuan Ren$^{1}$ \quad Guodongfang Zhao$^{2}$ \quad Bailan Feng$^{2}$  \quad Chao Ma$^{1}$\footnotemark[2]\\
    ${}^{1}$ MoE Key Lab of Artificial Intelligence, AI Institute, Shanghai Jiao Tong University \\
    ${}^{2}$ Central Research Institute, Huawei\\
    {\tt\small \{guoqing.wang, pin.tang, bunny\_renxiangxuan, chaoma\}@sjtu.edu.cn} \\
    {\tt\small \{zhaoguodongfang, fengbailan\}@huawei.com}
    }

\begin{document}
\twocolumn[{%
\renewcommand\twocolumn[1][]{#1}%
\maketitle
\vspace{-10mm}
\begin{center}
    \centering
    \includegraphics[width=1.0\linewidth]{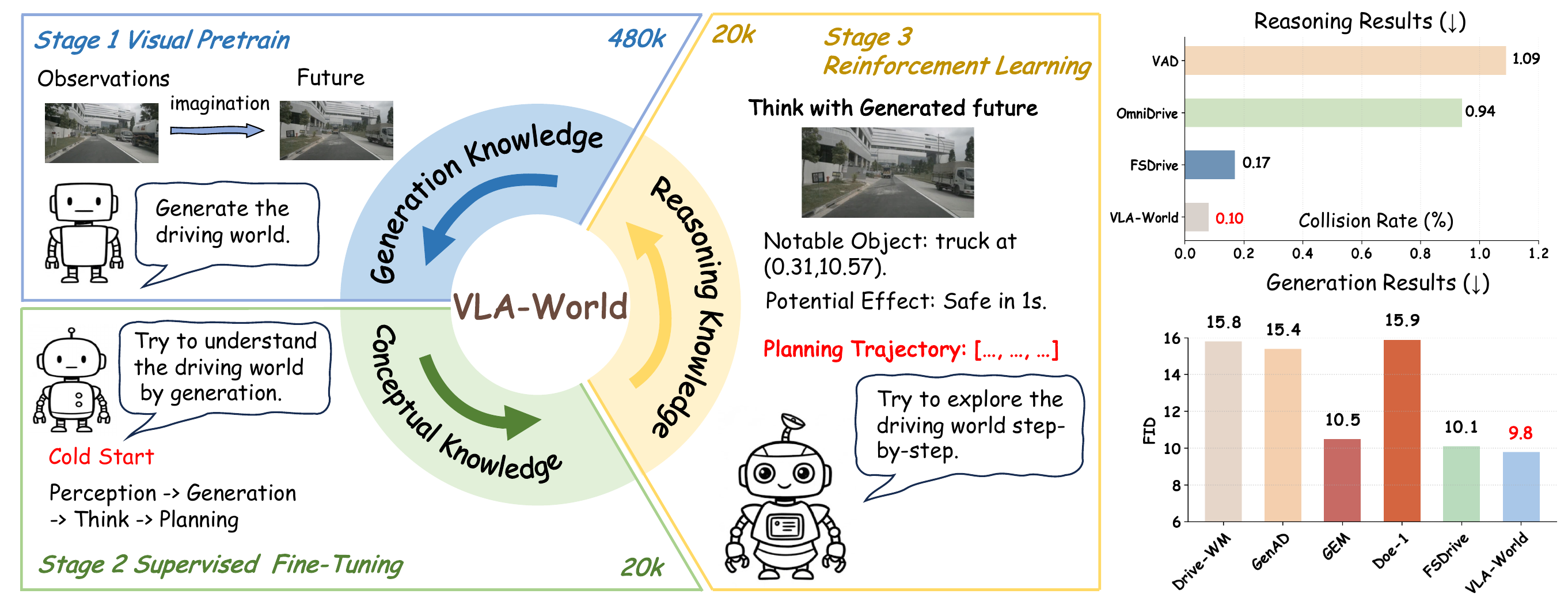}
    \captionof{figure}{ Visual overview of VLA-World. The model learns through three progressive stages. We first activate visual generation by predicting future frames from multi-view inputs to learn generation knowledge. Then, we fine-tune the model to link perception, future generation, and planning to learn driving conceptual knowledge. Finally, we use reinforcement learning to refine decisions through interaction with generated futures to explore reasoning knowledge. The results on the right show that VLA-World achieves both the lowest collision rate and FID score, highlighting its strengths in future generation and driving-oriented reasoning.
    }
\label{teaser}
\end{center}%
}]

\begingroup
\renewcommand{\thefootnote}{\textdagger}
\footnotetext{\scriptsize Corresponding author.}
\endgroup

\begin{abstract}
Vision-Language-Action (VLA) models have recently achieved notable progress in end-to-end autonomous driving by integrating perception, reasoning, and control within a unified multimodal framework. However, they often lack explicit modeling of temporal dynamics and global world consistency, which limits their foresight and safety. In contrast, world models can simulate plausible future scenes but generally struggle to reason about or evaluate the imagined future they generate.
In this work, we present VLA-World, a simple yet effective VLA world model that unifies predictive imagination with reflective reasoning to improve driving foresight. VLA-World first uses an action-derived feasible trajectory to guide the generation of the next-frame image, capturing rich spatial and temporal cues that describe how the surrounding environment evolves. The model then reasons over this self-generated future imagined frame to refine the predicted trajectory, achieving higher performance and better interpretability.
To support this pipeline, we curate nuScenes-GR-20K, a generative reasoning dataset derived from nuScenes, and employ a three-stage training strategy that includes pretraining, supervised fine-tuning, and reinforcement learning. Extensive experiments demonstrate that VLA-World consistently surpasses state-of-the-art VLA and world-model baselines on both planning and future-generation benchmarks. Project page: \url{https://vlaworld.github.io}

\end{abstract}

\section{Introduction}
\label{sec:intro}

\begin{figure}[!t]
    \centering
    \captionsetup{type=figure}
    \includegraphics[width=1.0\linewidth]{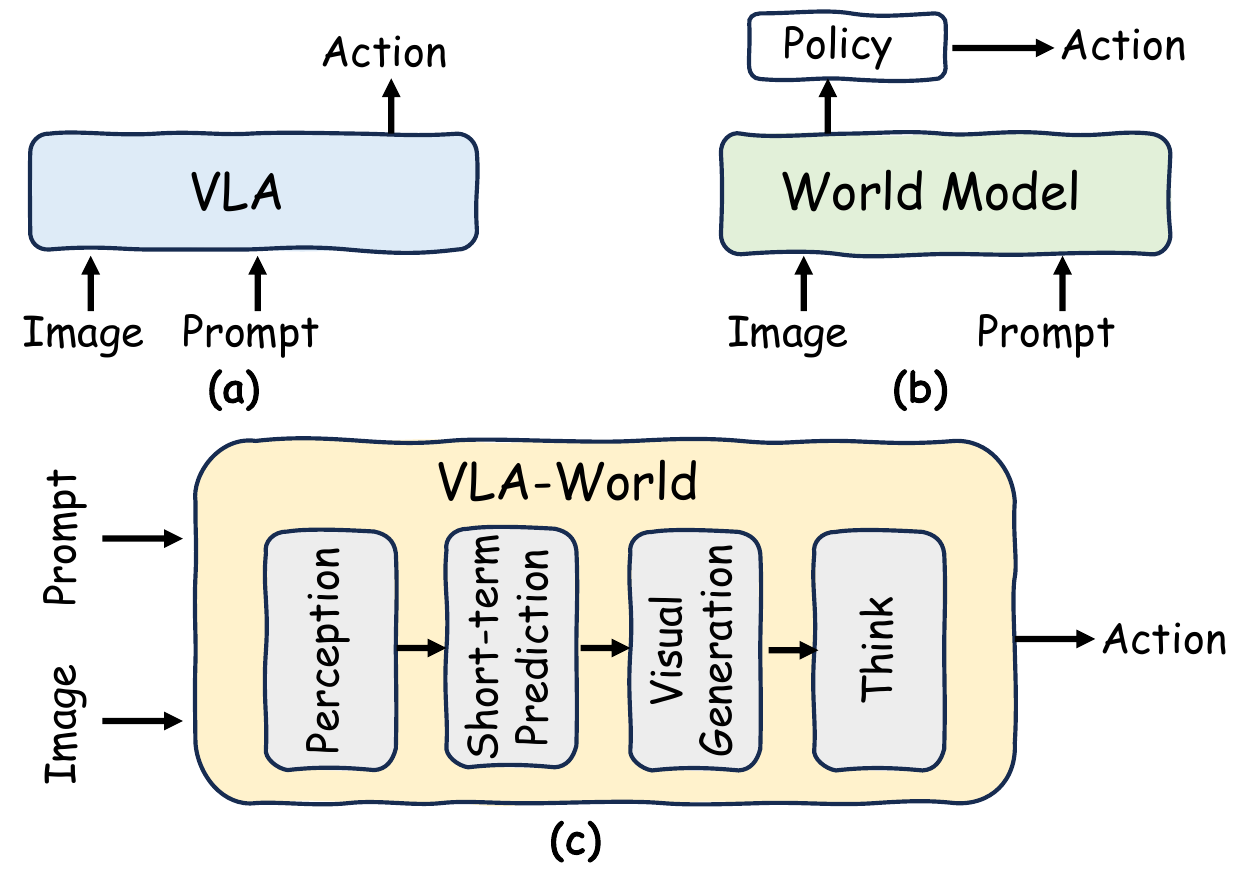}
    \vspace{-4mm}
    \caption{Comparison of the (a) VLA, (b) World Model, and (c) the proposed VLA-World paradigm.}
    \label{fig: paradigm}
    \vspace{-4mm}
\end{figure}


Recently, two major paradigms have gained attention in end-to-end autonomous driving: Vision-Language-Action (VLA) models~\cite{zhou2025autovla, zhou2025opendrivevla, alphadrive, ma2024dolphins, zhang2024chatscene, li2024llada, han2025dme, chi2025impromptu, arai2025covla} and World Models~\cite{wang2024driving, wang2023drivedreamer, wang2024worlddreamer, lu2025wovogen, jia2023adriver, min2024driveworld, Hu2023GAIA1AG}. Unlike traditional end-to-end pipelines~\cite{uniad, vad, vad2, chen2024end, chen2024ppad} that learn perception and control only from driving data, VLA models are built on powerful Vision-Language Models (VLMs)~\cite{achiam2023gpt4, bai2023qwenvl, bai2025qwen25vl, alayrac2022flamingo, liu2024llava, liu2024improved} pretrained on large vision-instruction datasets. By combining perception, language-based reasoning, and action generation in an autoregressive manner, they offer strong generalization and scalability.
World models focus on predicting how the environment will evolve by generating future frames. Using generative architectures, they learn latent spatiotemporal dynamics that allow the models to anticipate upcoming events and support more informed decision-making.

However, both paradigms face fundamental limitations. Existing VLA models~\cite{zhou2025autovla, zhou2025opendrivevla, tian2024drivevlm, alphadrive} inherently lack explicitly spatiotemporal modeling of other dynamic agents in the driving scenes, making it difficult to predict the evolution of complex scenarios, a capability that is essential for safe and proactive driving. In contrast, world models typically rely on large-scale visual data to learn a prior distribution and then sample from it, without capturing the underlying causal relationships of the world effectively. As a result, they tend to \textit{simulate} the world rather than truly \textit{understand} it. 

To address these limitations, recent works~\cite{wu2025janus, chen2025janus, xie2024show, cen2025worldvla, zeng2025futuresightdrive, zhang2025dreamvla} have begun exploring unified architectures that integrate generation and understanding, using generation as a bridge to enhance latent representations.
Building upon these works, we argue that an ideal paradigm for end-to-end autonomous driving should merge the spatiotemporal modeling strength of world models with the reasoning ability of VLA models. Such a pipeline would not only envision how the scene will evolve but also interpret and reflect on those imagined futures, much as human drivers do.
A vivid driving example illustrates this intuition. When cruising on an open road, a human driver relies on quick, intuitive imagination, like world models, to predict the next few moments without conscious effort. But if a pedestrian suddenly steps into the lane, the driver immediately shifts into reflective reasoning: the mind simulates what would happen if the car kept moving at the same speed, evaluates the outcome, and then overrides the initial impulse to continue forward.


We introduce VLA-World, a simple yet effective Vision-Language-Action World Model that not only generates short-term future frames conditioned on predicted short-term trajectories but also reasons over these imagined futures to assess potential risks. This enables more informed decisions and safer trajectory planning. Our key insight is that short-term predicted futures naturally encode rich spatiotemporal information about how the scene will evolve, capturing both ego motion and the behaviors of surrounding agents, which are essential for reliable driving reasoning.
As illustrated in Figure~\ref{fig: paradigm}, VLA-World follows a multi-step pipeline that includes perception, short-term prediction, generation, reasoning, and planning. VLA-World first perceives the environment by detecting relevant traffic participants and estimating distances to road boundaries. It then predicts the ego trajectory and driving direction for the next 0.5 seconds and generates the corresponding future frame based on these predictions. The model reasons over this generated future image to identify important agents and potential risks that may emerge. Finally, it outputs the appropriate driving decision along with the long-term trajectory. By integrating imagination and reasoning within one framework, VLA-World can both anticipate and reflect on upcoming events, resulting in more human-like and safety-aware driving behavior.


To support this pipeline, we curate a dataset from nuScenes~\cite{caesar2020nuscenes}, named nuScenes-GR-20K, specifically designed for generating future frames and reasoning conditioned on them. We further introduce a three-stage training strategy to fully explore the reasoning capability, as shown in Figure~\ref{teaser}. The process includes: (1) pretraining on large image–instruction datasets to activate visual generation knowledge, (2) supervised fine-tuning (SFT) on a multi-task mixed dataset to learn driving-related conceptual knowledge, and (3) reinforcement learning (RL) with Group Relative Policy Optimization (GRPO) to explore human-like reasoning knowledge. This three-stage training pipeline maintains end-to-end policy consistency and enables joint optimization across all components, from future generation to reasoning and planning. Extensive experiments show that VLA-World significantly outperforms state-of-the-art VLA and world model baselines on both generation and reasoning benchmarks, highlighting the effectiveness and versatility of our approach. In summary, our contributions are as follows:

\begin{itemize}
  \item We introduce a simple yet effective VLA world model for autonomous driving that unifies predictive imagination and reflective reasoning in a single framework.
  \item We curate nuScenes-GR-20K, a dataset for generation and reasoning, and propose a three-stage training strategy to fully unleash the intelligence of VLA-World.
  \item We thoroughly evaluate VLA-World and show that it achieves strong performance on both future-frame generation and planning benchmarks, outperforming previous VLA and world models.
\end{itemize}

\section{Related Work}
\label{sec:related}

\subsection{Vision-Language-Action Models}


Vision-Language-Action (VLA) models~\cite{zhou2025autovla, zhou2025opendrivevla, tian2024drivevlm, alphadrive, zheng2025driveagent, mao2023language, drivegpt4, cen2025worldvla, qian2025agentthink, nie2024reason2drive, brohan2022rt, zitkovich2023rt, belkhale2024rt, black2024pi_0, driess2023palm} have achieved notable progress in robotics and autonomous driving, driven by the rapid advances of multimodal large language models (MLLMs)~\cite{bai2023qwenvl, wang2024qwen2vl, bai2025qwen25vl, liu2024llava, li2022blip, li2023blip2, touvron2023llama, liu2024improved, ren2025grounding}.
Several works~\cite{drivegpt4, shao2024lmdrive, tian2024drivevlm, drivemlm} for autonomous driving leverage pretrained LLMs to generate driving actions accompanied by textual rationales, thus can enhance the reasoning and interpretability of the model. DriveMoE~\cite{drivemoe} introduces a Mixture of Experts framework that dynamically selects specialized networks to better handle diverse and complex scenarios. OmniDrive~\cite{wang2025omnidrive} presents a unified LLM-agent system that supports 3D perception, reasoning, and planning through a query-based 3D vision-language architecture and a new counterfactual benchmark. Inspired by the reasoning style of DeepSeek-R1~\cite{guo2025deepseek}, several methods~\cite{zheng2025driveagent, yuan2025autodrive, alphadrive} incorporate GRPO-based reinforcement learning~\cite{grpo} to strengthen reasoning and self-reflection in VLA models.
Despite these advances, current VLA approaches still have limited temporal understanding and weak world consistency, as they often map observations directly to actions without modeling how the environment evolves over time.

\subsection{World Models for Autonomous Driving}
Most existing world models~\cite{wang2023drivedreamer, wang2024driving, brooks2024video, zheng2024genad, zheng2024doe, zheng2024occworld, zheng2025world4drive, yang2024generalized, wu2023mars, xu2025occ, yang2025resim, yang2025driving, li2024uniscene, li2025omninwm} for autonomous driving focus on generating driving-compliant videos from past image sequences and current actions. A pioneering work is DriveDreamer~\cite{wang2023drivedreamer}, which uses a diffusion-based framework to generate realistic future driving videos and predict subsequent actions. 
DrivingWorld~\cite{hu2024drivingworld} introduces a GPT-style world model for autonomous driving, featuring several spatial-temporal fusion mechanisms, which enable effective modeling of both spatial and temporal dynamics, facilitating high-fidelity, long-duration video generation.
To maintain multi-view consistency due to the absence of an effective 3D spatial representation, OccWorld~\cite{zheng2024occworld} leverages the past 3D occupancy observations to generate future 3D occupancy maps. Most existing world models focus on generating temporally consistent future scenes but lack explicit reasoning or action awareness, limiting their interpretability and decision reliability. Recently, FSDrive~\cite{zeng2025futuresightdrive} introduces a spatiotemporal Chain-of-Thought (CoT)~\cite{wei2022chain, sarkar2025reasoning} based on Qwen2-VL~\cite{yang2024qwen2} model that \textit{thinks visually} by generating a future image frame as an intermediate reasoning step. Different from FSDrive, our proposed VLA-World integrates the predictive imagination of world models with the reflective reasoning of VLA frameworks, enabling both accurate foresight and cognitively grounded decision-making for autonomous driving.

\begin{figure*}[!t]
    \centering
    \captionsetup{type=figure}
    \includegraphics[width=1.0\linewidth]{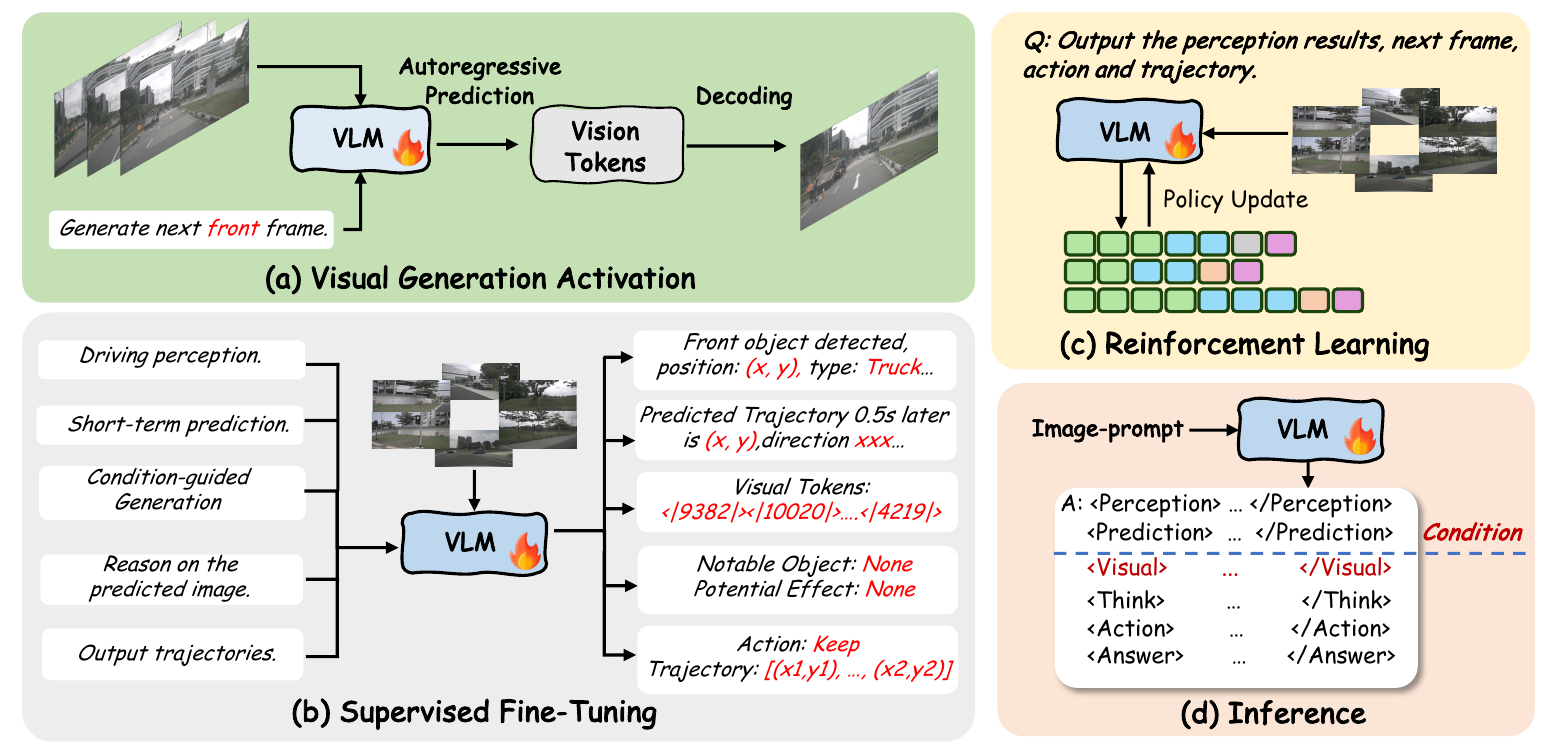}
    \caption{The illustration of the three-stage training and inference pipeline of VLA-World. Our training pipeline consists of three key stages: (a) visual pretrain to activate generation capability, (b) supervised fine-tuning to seed conceptual knowledge to the model, and (c) reinforcement learning with GRPO to explore higher performance like humans.}
    \label{fig: method}
\end{figure*}

\section{Proposed Method}
In this section, we first introduce the preliminaries on VLA and world models. Then, we present the overview of VLA-World. Finally, we introduce the three training stages in order, shown in Figure~\ref{fig: method}: (1) visual pretraining, (2) supervised fine-tuning, and (3) reinforcement learning.

\subsection{Preliminaries: VLA vs. World Models}
We consider an agent that drives in discrete time $t=\{1,2,\dots\}$ with a fixed step $\Delta t$ (e.g., 0.5 $\mathrm{s}$). At time $t$, the driving agent receives multi-view visual observations and its ego status $o_t = \{I_t^{1:K}, S_t\}$, where $I_t^{k}\in\mathbb{R}^{H\times W\times 3}$ denote the input images from camera $k\in \{1,\dots,K\}$ and $S_t\in\mathbb{R}^{d_s}$ includes ego-velocity, acceleration, yaw rate, and other CAN signals. The agent is conditioned on a mission goal $g$ (e.g., left/right/forward). We denote the future waypoint trajectory in an ego-centric BEV coordinate as
\begin{equation}
    \tau_{t:t+H} = \{ p_{t+1}, p_{t+2}, \dots, p_{t+H} \}, \quad p_{t+h}\in\mathbb{R}^2,
\end{equation}
and the next-frame image (for a chosen camera or view) as $x_{t+1}$. The action space can be low-level controls $a_t$ (throttle, brake, steering) or a trajectory $\tau_{t:t+H}$.

\noindent\textbf{VLA Models.} The VLA models learn a direct mapping from history observations and goals to trajectories as $\pi_\theta(\tau_{t:t+H}\mid o_{1:t}, g)$. This paradigm unifies perception, reasoning, and planning within a large language or multimodal language model framework. Similar to a one-stage end-to-end autonomous driving pipeline, it can be trained purely on large-scale trajectory data, making the overall optimization highly concise and efficient. However, VLA lacks explicit modeling of temporal dynamics and world consistency, which causes the model to focus mainly on the ego vehicle while neglecting the motion modeling of other traffic participants, an essential component that must be captured in end-to-end autonomous driving pipelines~\cite{uniad, vad, vad2}.

\noindent\textbf{World Models.} World models aim to capture the latent dynamics of the surrounding environment by predicting how the world evolves under actions. Formally, it learns a transition distribution $p_\psi(w_{t+1}\mid w_t, a_t)$, where $w_t$ is a latent state summarizing past observations $o_{1:t}$, and $a_t$ is the action taken by the agent. The latent state is typically obtained through an encoder, and the model can reconstruct or imagine future observations through a decoder. Intuitively, a world model serves as an internal simulator that allows an agent to \textit{dream} plausible futures, visualizing potential outcomes of its actions without physically interacting with the environment. By repeatedly rolling out $\{w_{t+1}, w_{t+2}, \ldots\}$, the model can perform long-horizon prediction and reasoning in latent space. While conventional world models are strong at temporal prediction and future imagination, they usually lack reflective reasoning, meaning they can simulate what may happen but cannot assess whether those imagined futures are safe, feasible, or desirable.

\subsection{Vision-Language-Action World Model}
VLA-World is a unified framework that combines the strengths of VLA and world models to improve decision-making in autonomous driving. It harnesses the predictive imagination of world models to simulate future scenarios and refines these predictions through reflective reasoning, enabling more accurate, interpretable, and safety-aware driving decisions.
We formulate this paradigm as
\begin{equation}
    \begin{split}
        p(\tau_{t:t+H}, &x_{t+1} \mid o_{1:t}, g) = \\ &\underbrace{p(\tau_{t:t+H} \mid o_{1:t}, g)}_{\text{decision / policy}}
        \cdot \underbrace{p(x_{t+1} \mid o_{1:t}, \tau_{t+1})}_{\text{imagination / world model}},
    \end{split}
\end{equation}
where $\tau_{t+1}$ conditions the near-future evolution. Pure VLA models focus on the left factor, and pure world models focus on the right factor. For safe and interpretable driving, reflective thinking is needed: after imagining $\hat{x}_{t+1}$, the policy should query future evidence and revise its plan accordingly. 

Given a sequence of observations $o_{1:t}$ and a mission goal $g$, VLA-World first predicts an initial future trajectory $\hat{\tau}_{t:t+1}$. Conditioned on this predicted trajectory and the past observations, the model then imagines the expected visual observation at the next time step:
\begin{equation}
    \hat{x}_{t+1} \sim p_\psi(x_{t+1} \mid o_{1:t}, \hat{\tau}_{t:t+1}).
\end{equation}
The generated image $\hat{x}_{t+1}$ represents the anticipated near-future view under its current plan. Rather than treating visual generation as an auxiliary output, we use this imagined future as an explicit cue for reflective reasoning:
\begin{equation}
    \tilde{\tau}_{t:t+H} = f_{\mathrm{ref}}\!\left(o_{1:t}, \hat{x}_{t+1}, \hat{\tau}_{t:t+1}\right),
\end{equation}
where $f_{\mathrm{ref}}$ denotes the reflective reasoning module. This refinement step preserves the intent of the initial prediction while correcting decisions that are unsafe or inconsistent with the self-generated future. The final trajectory $\tilde{\tau}_{t:t+H}$ therefore reflects both the model's predicted dynamics and its reasoning over the imagined future scene.

\noindent\textbf{Intuitive Insight.} Our core insight enables \textit{thinking and reflection through future images generated from short-term predicted trajectories}. First, the model produces an intuitive short-term plan that samples the high-dimensional future into a reasonable and trustworthy space. It then visualizes the intuitive outcome on its sketchpad, which contains rich spatiotemporal cues. Most importantly, the model explicitly reflects on the content of its own generated images to identify potential risks that intuition might have overlooked. This closed loop from simulation to reflection allows the model to first act on intuition to find a path and then thoughtfully evaluate its consequences to seek the optimal solution.

\subsection{Visual Pretraining of VLA-World}

Following the alignment strategy of FSDrive~\cite{zeng2025futuresightdrive}, our visual pretraining stage aims to activate both the visual understanding and visual generation abilities of VLA-World, allowing the model to interpret complex driving scenes and imagine their short-term evolution. Unlike FSDrive, which generates future frames only for the front view, our pretraining stage explicitly enforces multi-view consistency, enabling the model to later produce coherent future images from any camera viewpoint required in the SFT and RL stages.

Formally, given a multi-view image set $I=\{I^{k}_t\}_{k=1}^{K}$ and an instruction $L$ describing the desired view or driving intent (e.g., generate CAM\_FRONT\_LEFT 0.5 s later), the model learns to predict the next visual token sequence $Q_{t+1}^{k}$ through autoregressive next-token prediction:
\begin{equation}
    P(Q_{t+1}^{k}) = \prod_{i=1}^{N} P_\theta(q_i^{k} \mid q_{<i}^{k}, h_t, L),
\end{equation}
where $h_t = f_\phi(I_t, S_t)$ encodes the current multi-view observations and ego state $S_t$, and $q_i^{k}$ denotes the $i$-th discrete token of the VQGAN~\cite{esser2021taming, van2017neural} codebook for camera $k$. The generated tokens can be decoded into future images $\hat{I}_{t+1}^{k}$ using the visual tokenizer of VQGAN.

This formulation allows the model to learn how each camera view evolves based on motion and control cues, giving it a unified spatiotemporal prior across all perspectives. This design ensures that in downstream tasks, when the planner predicts a short-term trajectory and requests the corresponding future view (such as turning left, turning right, or going forward), the pretrained generator can produce consistent and physically plausible images from any viewpoint. In this way, we extend FSDrive into a multi-view, goal-conditioned world model, creating a strong foundation for reflective reasoning and safety-aware planning in later stages.

\subsection{Supervised Fine-Tuning of VLA-World}
We perform supervised fine-tuning to seed driving conceptual knowledge into the base model through imitation learning, following the \textit{generation-to-think} paradigm. To achieve this, we train VLA-World on a comprehensive multi-task mixed dataset that is carefully designed to cover several essential learning objectives.

\noindent\textbf{Perception.} 
The perception module acts as the visual grounding stage of VLA-World, converting raw multi-view inputs into structured spatial and semantic representations that support later short-term prediction and reasoning. Using six camera-view images together with ego status, it detects surrounding dynamic agents such as vehicles and pedestrians, estimates their 3D positions, possible motion paths, road-shoulder distances, and drivable-area boundaries. These outputs form a scene-level world state, a compact and interpretable summary that captures object categories, spatial layout, and motion cues, providing the foundation for ego-trajectory prediction and reflective reasoning to ensure safety and consistency in future scenarios.

\noindent\textbf{Short-term Prediction.}
The short-term prediction module transforms the current perception results and ego status into near-future estimates of how the world will evolve, providing the basis for trajectory planning and temporal consistency in VLA-World. Using the history of ego states, it predicts the next waypoint and driving direction at regular intervals, for example, every 0.5 seconds. By explicitly modeling short-horizon dynamics, this module supports the visual generation step that imagines the corresponding future frame and ensures that the predicted trajectories remain smooth, temporally coherent, and physically plausible.

\noindent\textbf{Condition-guided Generation.} The generation module serves as the imagination core of VLA-World, transforming the predicted trajectory and direction into a fixed number of visual tokens of the near future. Conditioned on the encoded scene context $o_{1:t}$  and the predicted waypoint $\hat{\tau}_{t+1}$, this module generates the next-frame image $\hat{x}_{t+1}$, effectively visualizing how the environment is expected to evolve if the planned trajectory were executed, expressed as a compact set of visual tokens that encode spatial layout, object motion, and lighting continuity. By bridging low-level perception and high-level reasoning, this module not only offers interpretable evidence for the agent’s future state but also supplies the reflective reasoning module with a concrete scene hypothesis to evaluate for safety and consistency. This imagination step thus enables VLA-World to couple action-conditioned prediction with visual foresight, forming the foundation for reflective refinement in subsequent stages.

\noindent\textbf{Thinking with Visual Tokens.}
The thinking module embodies the reflective reasoning process of VLA-World, bridging imagination and decision refinement through causal interpretation of the generated future. After the generation module produces the next-frame prediction $\hat{x}_{t+1}$, the reflective reasoning analyzes salient entities, motion cues, and potential interactions to assess environmental risks and behavioral implications. This reflective process transforms visual evidence into situational understanding, quantifying safety margins, anticipating conflicts, and validating trajectory feasibility. Functionally, the think module serves as the cognitive layer of VLA-World, enabling the agent not only to predict what will happen but also to reason about whether it should happen, thus providing the foundation for trajectory refinement and safety-aware action in the subsequent stage.

\noindent\textbf{Action and Trajectory Planning.} These two modules constitute the final outputs of VLA-World, transforming reflective understanding into concrete driving behavior. After the think module evaluates the safety and feasibility of the predicted scenario, the model determines the appropriate action policy or maneuver that aligns with both the mission goal and reflective reasoning outcomes. Then, we translate this high-level action into a sequence of explicit spatial waypoints $\tilde{\tau}_{t:t+H}$, representing the refined ego-trajectory at 0.5 s intervals over a 3s horizon.
Together, these stages close the \textit{perception–prediction-imagination–reflection–action} loop of VLA-World, ensuring that final decisions are both context-aware and future-consistent, executing maneuvers that are not only physically feasible but also reflectively validated for safety and goal alignment.

\subsection{Reinforcement Learning of VLA-World}


Building on the SFT-trained model, we further adopt GRPO~\cite{grpo} algorithm to strengthen the advanced reasoning and decision-making capabilities of VLA-World. As shown in Figure~\ref{fig: method}, this phase shifts the model from following predefined reasoning patterns to dynamically formulating optimal planning strategies through an iterative and self-correcting process.
For each input prompt, GRPO samples a diverse set of candidate responses from the current policy. Then, we carefully designed a collection of rule-based reward functions to assess the quality of these responses across the entire VLA-World pipeline, spanning perception, short-term prediction, visual generation, and planning:

\noindent\textbf{Format Reward ($R_{fmt}$).} 
This reward enforces a well-structured output format. The perception description should appear inside the \texttt{<Perception>} tag, the short-term trajectory and driving direction inside the \texttt{<Prediction>} tag, the generated visual tokens inside the \texttt{<Visual>} tag, the reasoning content inside the \texttt{<Think>} tag, and the final high-level action and three-second trajectory within the \texttt{<Action>} and \texttt{<Answer>} tags.

\noindent\textbf{Short-term Prediction Reward ($R_{pred}$).} 
This reward serves two purposes: (1) it encourages accurate prediction of the short-term trajectory and heading, which conditions future frame generation; and (2) it enforces consistency between the 0.5-second prediction and the refined long-term trajectory produced after reasoning.

\noindent\textbf{Visual Constrain Reward ($R_{vis}$).} 
This reward ensures that the number of generated visual tokens matches the required length for correct image reconstruction. In addition, every token must correspond to a valid entry in the visual codebook to guarantee a decodable and meaningful generated frame.

\noindent\textbf{Action Reward ($R_{act}$).} 
To assess the correctness of the predicted high-level action, we compute a reward derived from the F1 score, which provides a balanced evaluation of precision and recall relative to the ground truth action set.

\noindent\textbf{Trajectory Reward ($R_{traj}$).} 
This reward ensures that the predicted trajectory over the final three seconds is accurate at each interval, while also enforcing kinematic consistency. For example, for an agent moving smoothly, the changes in acceleration should remain very small.

The final reward is computed as a weighted combination of all the above components:
\begin{equation}
    R_{\text{all}} = \lambda_{\text{fmt}} \cdot R_{\text{fmt}}
+ \lambda_{\text{pred}} \cdot R_{\text{pred}}
+ \lambda_{\text{vis}} \cdot R_{\text{vis}}
+ \lambda_{\text{act}} \cdot R_{\text{act}}
+ \lambda_{\text{traj}} \cdot R_{\text{traj}} \, .
\end{equation}
Overall, this training stage guides VLA-World toward producing outputs that are structurally correct, short-term prediction reasonable, visually coherent, and behaviorally safe, ultimately enabling more reliable driving decisions.


\section{Experiments}
\begin{table*}[t]
    \centering
    \setlength{\tabcolsep}{1.2mm}
    \caption{End-to-end trajectory planning results on nuScenes~\cite{caesar2020nuscenes}. We evaluate L2 error and collision rate following the respective protocols of ST-P3~\cite{hu2022stp3} and UniAD~\cite{uniad}. The $*$ indicates the use of additional ego-state information. Results for VAD~\cite{vad} and UniAD~\cite{uniad} are taken from BEV-Planner~\cite{bev-planner}.}
    \vspace{-7pt}
    \resizebox{1.0\linewidth}{!}{
    \begin{tabular}{lcccccccc|cccccccc|c}
    \toprule
     \multirow{4}{*}{\textbf{Method}} & \multicolumn{8}{c}{\textbf{ST-P3 metrics}} & \multicolumn{8}{c}{\textbf{UniAD metrics}} & \multirow{4}{*}{{\textbf{LLM}}}  \\
    \cmidrule(lr){2-9} \cmidrule(lr){10-17}
    & \multicolumn{4}{c}{\textbf{L2 (m)} $\downarrow$} & \multicolumn{4}{c}{\textbf{Collision (\%)} $\downarrow$} & \multicolumn{4}{c}{\textbf{L2 (m)} $\downarrow$} & \multicolumn{4}{c}{\textbf{Collision (\%)} $\downarrow$} &   \\
    \cmidrule(lr){2-5} \cmidrule(lr){6-9} \cmidrule(lr){10-13} \cmidrule(lr){14-17}
    & 1s & 2s & 3s & \cellcolor{gray!20}Avg. & 1s & 2s & 3s & \cellcolor{gray!20}Avg. & 1s & 2s & 3s & \cellcolor{gray!20}Avg. & 1s & 2s & 3s & \cellcolor{gray!20}Avg. &  \\
    \midrule
    \multicolumn{18}{c}{\textbf{Non-Autoregressive methods}} \\
    \midrule
    ST-P3*~\pub{ECCV22}~\cite{hu2022stp3}& 1.33 & 2.11 & 2.90 &   \cellcolor{gray!30}2.11 & 0.23 & 0.62 & 1.27 &   \cellcolor{gray!30}0.71 & - & - & - &   - & - & - & - &   - & -  \\ 

     VAD~\pub{ICCV23}~\cite{vad} & 0.69 & 1.22 & 1.83 &   \cellcolor{gray!30}1.25 & 0.06 & 0.68 & 2.52 &   \cellcolor{gray!30}1.09 & - & - & - &   - & - & - & - &   - & -  \\
    VAD*~\pub{ICCV23}~\cite{vad} & 0.17 & 0.34 & 0.60 &   \cellcolor{gray!30}0.37 & 0.04 & 0.27 & 0.67 &   \cellcolor{gray!30}0.33 & - & - & - &   - & - & - & - &   - & -  \\
    
    UniAD~\pub{CVPR23}~\cite{uniad} & - & - & - & - & - & - & - & - & 0.59 & 1.01 & 1.48 &   \cellcolor{gray!30}1.03 & 0.16 & 0.51 & 1.64 &   \cellcolor{gray!30}0.77 & -  \\
    UniAD*~\pub{CVPR23}~\cite{uniad} & - & - & - & - & - & - & - & - & 0.20 & 0.42 & \textbf{0.75} &   \cellcolor{gray!30}0.46 & 0.02 & 0.25 & 0.84 &   \cellcolor{gray!30}0.37 & -  \\

    BEV-Planner~\pub{CVPR24}~\cite{bev-planner} & 0.30 & 0.52 & 0.83 &   \cellcolor{gray!30}0.55 & 0.10 & 0.37 & 1.30 &   \cellcolor{gray!30}0.59 & - & - & - &   - & - & - & - &   - & -  \\
    BEV-Planner*~\pub{CVPR24}~\cite{bev-planner} & 0.16 & 0.32 & 0.57 &   \cellcolor{gray!30}0.35 & \textbf{0.00} & 0.29 & 0.73 &   \cellcolor{gray!30}0.34 & - & - & - &   - & - & - & - &   - & -  \\

    PreWorld~\pub{ICLR25}~\cite{li2025semi} & - & - & - &   - & - & - & - & - & 0.49  & 1.22 & 2.32 & \cellcolor{gray!30}1.34 & 0.19 & 0.57 & 2.65 & \cellcolor{gray!30}1.14&- \\
    
    \midrule
    \multicolumn{18}{c}{\textbf{Autoregressive methods}} \\
    \midrule

     ELM~\pub{ECCV24}~\cite{zhou2024embodied}& - & - & - &   - & - & - & - & -&  0.34  & 1.23 & 2.57 & \cellcolor{gray!30}1.38 & 0.12 & 0.50 & 2.36 & \cellcolor{gray!30}0.99 & BLIP2-2.7B\\
    FeD*~\pub{CVPR24}~\cite{zhang2023coaching} & - & - & - &   - & - & - & - & -& 0.27  & 0.53 & 0.94 & \cellcolor{gray!30}0.58 & \textbf{0.00} & \textbf{0.04} & 0.52 & \cellcolor{gray!30}0.19 & LLaVA-7B\\

    OccWorld~\pub{ECCV24}~\cite{zheng2024occworld} & 0.39  & 0.73 & 1.18 & \cellcolor{gray!30}0.77 & 0.11 & 0.19 & 0.67 & \cellcolor{gray!30}0.32 & 0.52  & 1.27 & 2.41 & \cellcolor{gray!30}1.40 & 0.12 & 0.40 & 2.08 & \cellcolor{gray!30}0.87 &GPT3-like\\
    
    Doe-1~\pub{arxiv24}~\cite{zheng2024doe}  & 0.37  & 0.67 & 1.07 & \cellcolor{gray!30}0.70 & 0.02 & 0.14 & 0.47 & \cellcolor{gray!30}0.21& 0.50  & 1.18 & 2.11 & \cellcolor{gray!30}1.26 & 0.04 & 0.37 & 1.19 & \cellcolor{gray!30}0.53 &Lumina-mGPT-7B\\
    
    RDA-Driver*~\pub{ECCV24}~\cite{RDA-Driver}& 0.17 & 0.37 & 0.69 &   \cellcolor{gray!30}0.40 & 0.01 & 0.05 & 0.26 &   \cellcolor{gray!30}\textbf{0.10} & 0.23 & 0.73 & 1.54 &   \cellcolor{gray!30}0.80 & \textbf{0.00} & 0.13 & 0.83 &   \cellcolor{gray!30}0.32 & LLaVA-7B  \\
    
    EMMA*~\pub{arxiv24}~\cite{hwang2024emma} & 0.14 & 0.29 & 0.54 & \cellcolor{gray!30}0.32 & - & - & - &   - & - & - & - & - & - & - & - & - & Gemini 1.0 Nano-1 \\


    OmniDrive~\pub{CVPR25}~\cite{wang2025omnidrive} & 0.40  & 0.80 & 1.32 & \cellcolor{gray!30}0.84 & 0.04 & 0.46 & 2.32 & \cellcolor{gray!30}0.94 & - & - &  - & - & - & - & - & - & LLaVA-7B \\
    OmniDrive*~\pub{CVPR25}~\cite{wang2025omnidrive} & 0.14  & 0.29 & 0.55 & \cellcolor{gray!30}0.33 & \textbf{0.00} & 0.13 & 0.78 & \cellcolor{gray!30}0.30  & - & - & - & - & - & - & - & - & LLaVA-7B \\
    
    FSDrive~\pub{NeurIPS25}~\cite{zeng2025futuresightdrive} & 0.28  & 0.52 & 0.80 & \cellcolor{gray!30}0.53 & 0.06 & 0.13 & 0.32 & \cellcolor{gray!30}0.17 & 0.40  & 0.89 & 1.60 & \cellcolor{gray!30}0.96 & 0.07 & 0.12 & 1.02 & \cellcolor{gray!30}0.40& Qwen2-VL-2B  \\
    FSDrive*~\pub{NeurIPS25}~\cite{zeng2025futuresightdrive} & 0.14  & 0.25 & 0.46 & \cellcolor{gray!30}0.28 & 0.03 & 0.06 & \textbf{0.21} & \cellcolor{gray!30}0.10 & 0.18  & 0.39 & 0.77 & \cellcolor{gray!30}0.45 & \textbf{0.00} & 0.06 & 0.42 & \cellcolor{gray!30}0.16 & Qwen2-VL-2B  \\
    \midrule

    \textbf{VLA-World (ours)} & 0.11 & 0.27 & 0.52 & \cellcolor{gray!30}0.30 & 0.00 & \textbf{0.03} & 0.26 & \cellcolor{gray!30}0.10  &  0.38 & 0.74 & 1.38 & \cellcolor{gray!30}0.83 & 0.02 & 0.08 & 0.36 & \cellcolor{gray!30}0.16& Qwen2-VL-2B  \\
    \textbf{VLA-World* (ours)} & \textbf{0.10} & \textbf{0.24} & \textbf{0.45} & \cellcolor{gray!30}\textbf{0.26} & 0.02 & 0.05 & \textbf{0.18} & \cellcolor{gray!30}\textbf{0.08} & \textbf{0.10} & \textbf{0.35} & 0.80 & \cellcolor{gray!30}\textbf{0.42} & 0.01 & 0.05 & \textbf{0.30} & \cellcolor{gray!30}\textbf{0.12}& Qwen2-VL-2B  \\

    \bottomrule
    \end{tabular}
    }
    \label{tab:plan}
\end{table*}


\subsection{Experimental Setup}

\textbf{Datasets and Metrics.} We conduct experiments on nuScenes dataset~\cite{caesar2020nuscenes} following the traditional end-to-end methods~\cite{uniad, vad, vad2}, VLA~\cite{wang2025omnidrive, zeng2025futuresightdrive, hwang2024emma} and world models~\cite{wang2023drivedreamer,kim2021drivegan,wang2024driving, zheng2024doe}. We curate a nuScenes-GR-20K dataset, including 20K samples for generating future frames and reasoning conditioned on them for SFT and RL stages. We evaluate trajectory planning performance using L2 displacement error and collision rate, following established protocols in prior studies~\cite{vad, vad2, hu2022stp3, wang2025omnidrive, zeng2025futuresightdrive}. In addition, consistent with prior works for generation~\cite{wang2023drivedreamer, wang2024driving}, we employ the Fréchet Inception Distance (FID) to evaluate the visual quality of generated future frames. More details are listed in the supplementary materials.

\noindent\textbf{Implementation Details.} We initialize our model with Qwen2-VL-2B~\cite{wang2024qwen2vl} following FSDrive~\cite{zeng2025futuresightdrive}. All training is conducted on $8\times80$ GB GPUs using the PyTorch framework. 
During the pretraining stage, the model is trained for 30 epochs using AdamW with an initial learning rate of $5 \times 10^{-4}$, a per-device batch size of 16. For supervised fine-tuning, we train the model for 12 epochs with AdamW and an initial learning rate of $1 \times 10^{-4}$.
Starting from the SFT checkpoint, the model undergoes an additional optimization phase using the GRPO for one epoch. The policy is trained with a learning rate of $1 \times 10^{-6}$ and a global batch size of 16. 
For each prompt, we sample 8 candidate responses to estimate the policy gradient. More implementation details can be found in supplementary materials.

\begin{table*}[!t]\small
\begin{center}
\setlength{\tabcolsep}{1.4mm}
\caption{Comparison of future frame generation results with different generative models evaluated by FID ($\downarrow$) on the nuScenes~\cite{caesar2020nuscenes} dataset.
}
\vspace{-7pt}
\resizebox{0.9\linewidth}{!}{ 
\begin{tabular}{l|ccccccc|c}
\toprule
\multirow{2}{*}{\textbf{Method}}&DriveGAN&DriveDreamer&Drive-WM&GenAD & GEM &Doe-1&FSDrive& \multirow{2}{*}{\textbf{VLA-World}}   \\
&~\pub{CVPR21~\cite{kim2021drivegan}}&~\pub{ECCV24~\cite{wang2023drivedreamer}}&~\pub{CVPR24~\cite{wang2024driving}}&~\pub{CVPR24~\cite{yang2024generalized}} &~\pub{CVPR25~\cite{hassan2025gem}}&~\pub{arxiv24~\cite{zheng2024doe}}&~\pub{NeurIPS25~\cite{zeng2025futuresightdrive}}&    \\
\midrule
\textbf{Type}&GAN&Diffusion &Diffusion&Diffusion&Diffusion&Autoregressive &Autoregressive&Autoregressive  \\
\textbf{Resolution}  & 256$\times$256 & 128$\times$192  & 192$\times$384 & 256$\times$448 & 576$\times$1024& 384$\times$672&128$\times$192&128$\times$192\\
\midrule
\textbf{FID} $\downarrow$& 73.4& 52.6& 15.8& 15.4&10.5 &15.9& 10.1& \textbf{9.8}\\
\bottomrule
\end{tabular} 
\label{tab:fid}
}
\vspace{-10pt}
\end{center}
\end{table*}


\subsection{Main Results}

\noindent\textbf{End-to-End Trajectory Planning.}
We evaluate our method with both non-autoregressive and autoregressive baselines using ST-P3 and UniAD metrics on nuScenes. VLA-World achieves the best overall performance on both benchmarks, showing the lowest average L2 error and collision rate among autoregressive methods. 
Compared with non-autoregressive planners such as BEV-Planner~\cite{bev-planner} and UniAD~\cite{uniad}, VLA-World maintains better performance while producing more stable long-horizon predictions, benefiting from its generative world-model architecture. Unlike FSDrive, which directly regress future waypoints without evaluating physical feasibility, VLA-World integrates short-term imagination with reflective correction, leading to stronger foresight and reduced temporal drift. The improvements are most pronounced at 3-second horizons, where traditional VLA models tend to accumulate error. Overall, these results confirm that the proposed framework, combining action-conditioned future frame generation with reflective trajectory refinement, yields significant gains in both safety and trajectory fidelity, establishing VLA-World as a state-of-the-art method for autonomous driving.

\begin{figure}[!t]
    \centering
    \captionsetup{type=figure}
    \includegraphics[width=1.0\linewidth]{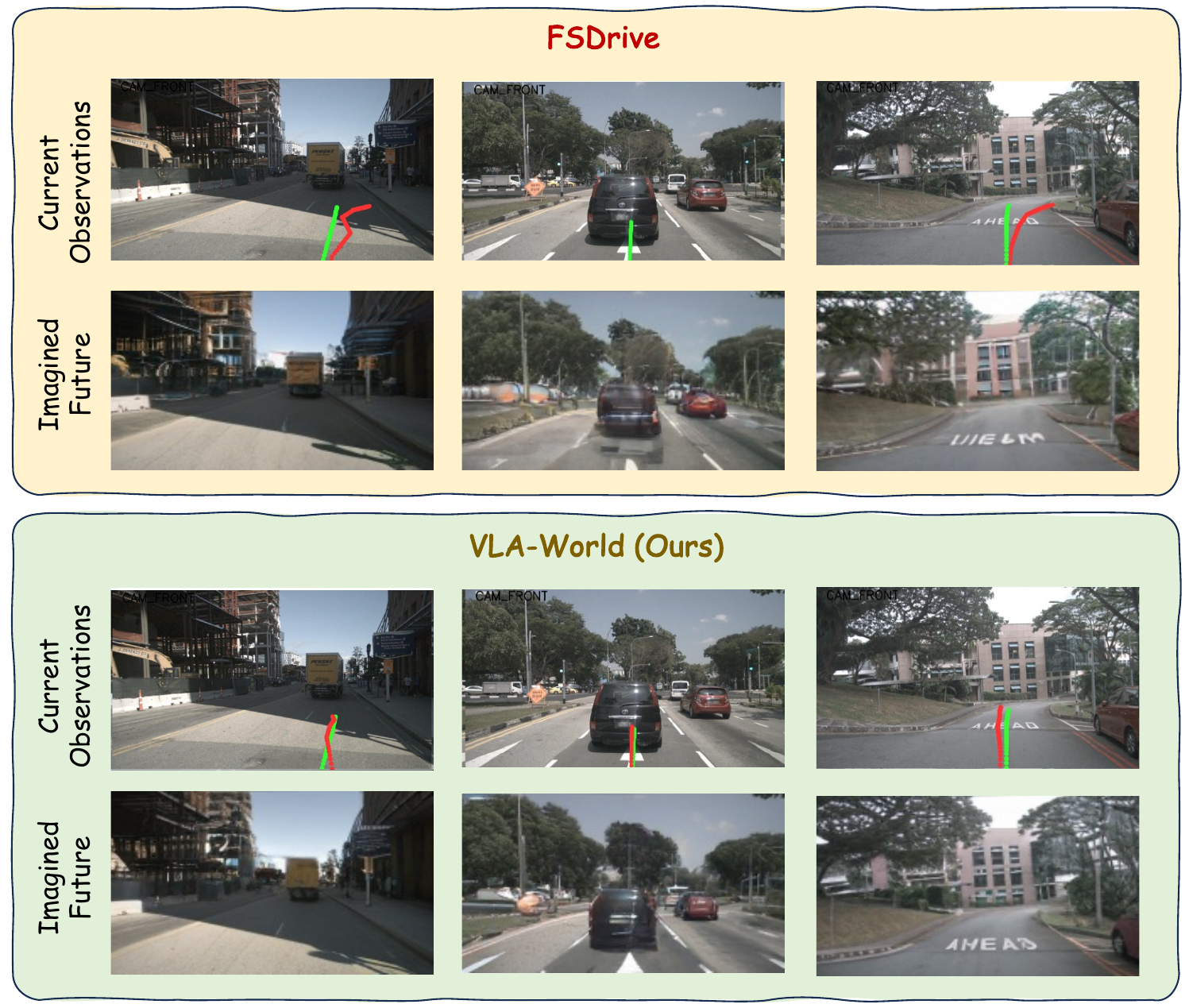}
    \vspace{-7pt}
    \caption{Visualization of our VLA-World compared with the SOTA FSDrive~\cite{zeng2025futuresightdrive}. The top row shows the current scene, and the bottom row shows the generated scene for the next 0.5 seconds. The red trajectory is the model prediction, and the green trajectory is the ground truth. Zoom in for a better view.}
    \vspace{-7pt}
    \label{fig: vis}
\end{figure}

\begin{table}
\centering
\caption{Action prediction performance on the nuScenes dataset measured by F1 score (\%). ${\text{\textdagger}}$ denotes models trained on nuScenes.}
\setlength{\tabcolsep}{3pt}
\renewcommand{\arraystretch}{1.0}
\vspace{-7pt}
\resizebox{1.0\linewidth}{!}{
        \begin{tabular}{l|ccc|cccc}
        \toprule
        \multirow{2}{*}{Method} & \multicolumn{3}{c|}{Lateral $($F1$)$ $\uparrow$} & \multicolumn{4}{c}{Longitudinal $($F1$)$ $\uparrow$} \\
        & forward & left & right & keep & acc. & dec. & stop \\
        \midrule
        Qwen2-VL-2B  & 62.43 & 22.75 & 28.65 & 40.70 & 50.23 & 49.21 & 41.04  \\
        \midrule
        Qwen2-VL-2B${\text{\textdagger}}$ & 92.60 & 61.78 & 66.52 & 56.42 & 74.32 & 76.10 & 74.85 \\
        \midrule
        \cellcolor[gray]{.9}VLA-World & \textbf{95.88} & \textbf{74.22} & \textbf{75.06} & \textbf{60.98} & \textbf{81.42} & \textbf{80.04} & \textbf{81.24} \\
        \bottomrule
        \end{tabular}
    } 
    \vspace{-10pt} 
 \label{table:action}
\end{table}

\noindent\textbf{Evaluation of Action.}
Table~\ref{table:action} shows the action prediction results of VLA-World across both lateral and longitudinal categories, evaluated using the F1-score. Our model achieves clear and consistent improvements in all action types, demonstrating its strong capability in learning goal-conditioned control. Compared with the base Qwen2-VL-2B backbone, the model fine-tuned on nuScenes (Qwen2-VL-2B†) shows a substantial performance boost, particularly for turning behaviors where the left category improves from 22.75\% to 61.78\% in F1-score. Our VLA-World delivers the best results across every metric, achieving 95.88\% for forward, 74.22\% for left, and 75.06\% for right. These improvements highlight the effectiveness of our VLA-World pipeline. It learns to reason about the consequences of its own actions rather than only imitating labels by reinforcement learning.

\noindent\textbf{Quantitative Results of Generation.} 
Although the ultimate goal of VLA-World is trajectory planning, the generation of the next frame serves only as an intermediate step to support subsequent reasoning. Nevertheless, as shown in Table~\ref{tab:fid}, we evaluate the visual quality using the FID metric. Our method achieves competitive performance compared with dedicated diffusion models~\cite{wang2023drivedreamer, wang2024driving, yang2024generalized}. Moreover, compared with Doe-1~\cite{zheng2024doe} and FSDrive~\cite{zeng2025futuresightdrive}, our approach still demonstrates a significant advantage, indicating that even with limited data, it can effectively unleash the visual generation capability of multimodal large language models. 

\noindent\textbf{Visualization Results.}
We present a qualitative comparison of trajectory prediction and future-frame generation between FSDrive~\cite{zeng2025futuresightdrive} and our VLA-World. As shown in Figure~\ref{fig: vis}, VLA-World produces noticeably sharper and more coherent future frames, benefiting from visual generation that is explicitly conditioned on the predicted trajectory and driving direction. Furthermore, VLA-World can perform reflective reasoning over the imagined temporal cues, enabling it to refine its motion forecasts, resulting in future trajectories that are more precise compared with FSDrive.


\subsection{Ablation Study}
\noindent\textbf{Effectiveness of training strategy.} 
We conduct ablation studies on the training strategy of VLA-World. \textit{w/o. P.T.}, \textit{w/o. SFT}, and \textit{w/o. RL} denote variants of the proposed method with the pretraining, supervised fine-tuning (cold start), and reinforcement learning stages removed, respectively. As shown in Table~\ref{tab:abl} (a), we observe that although each stage contributes to overall performance improvement, the variant trained with SFT outperforms the one trained with RL. This indicates that reinforcement learning without cold-start supervision struggles to effectively navigate the large search space of our structured, multi-step reasoning task. Therefore, SFT is crucial for instilling coherent policies and a fundamental understanding of causal chains. The pretraining stage also provides additional gains by enhancing the model’s spatiotemporal understanding of driving environments through future frame generation. The superior performance of the complete VLA-World model confirms the importance of our hybrid training design: pretraining aligns generation and understanding, SFT builds the essential knowledge foundation, and RL further refines the policy to achieve optimal performance.

\noindent\textbf{Effectiveness of data pipeline.} To verify the effectiveness of each step in our data pipeline, we perform ablation studies on these three components. As shown in Table~\ref{tab:abl} (b), all components contribute positively to overall performance. In particular, perception and reasoning have a greater impact compared to visual generation. We attribute this to the fact that visual generation involves producing a large number of tokens, which dominate the gradient updates during optimization, potentially limiting the exploration of the upper bound of model performance.

\noindent\textbf{Effectiveness of different rewards.} We also conduct ablation studies on the contribution of different rewards in the reinforcement learning stage. As shown in Table~\ref{tab:abl} (c), all reward terms have a positive effect on policy optimization, with the trajectory and action rewards contributing the most. This indicates that reinforcement learning can directly optimize the planning process in an end-to-end manner.

\begin{table}[]
    \centering
    \footnotesize
    \caption{Ablation studies of trajectory planning L2 errors (ST-P3) on nuScenes to validate each proposed component.}
    \setlength{\tabcolsep}{7pt}
    \renewcommand{\arraystretch}{0.85}
    \vspace{-7pt}
    \resizebox{1.0\linewidth}{!}{
    \resizebox{\linewidth}{!}{
    \begin{tabular}{l|cccccc}
\toprule 
&\multirow{2}{*}{\textbf{Method}} & \multicolumn{4}{c}{\textbf{L2 Error (m) $\downarrow$}} \\
\cmidrule(lr){3-6}
 & & 1s & 2s & 3s & Avg. \\
\midrule
\multirow{3}{*}{(a)} & w/o. P.T. & 0.35 & 0.56 & 0.81 & 0.57\\
&w/o. SFT & 0.35 & 0.79 & 1.40 & 0.85  \\
&w/o. RL  & 0.43 & 0.70 & 1.01 & 0.71   \\
\midrule
\multirow{3}{*}{(b)} & w/o. Perception & 0.42 & 0.73 & 1.09 & 0.75 \\
&w/o. Generation &  0.41 & 0.67 & 0.96 & 0.68 \\
&w/o. Reasoning & 0.50 & 0.83 & 1.22 & 0.85 \\
\midrule
\multirow{5}{*}{(c)}&w/o. $R_{\text{pred}}$ & 0.17 & 0.37 & 0.69 & 0.41 \\
&w/o. $R_{\text{vis}}$ & 0.20 & 0.40 & 0.67 & 0.42 \\
&w/o. $R_{\text{act}}$ & 0.40 & 0.54 & 0.92 & 0.62 \\
&w/o. $R_{\text{traj}}$ & 0.46 & 0.75 & 0.96 & 0.72 \\
\midrule
(d) & \cellcolor[gray]{.9}{VLA-World} & \textbf{0.11} & \textbf{0.27} & \textbf{0.52} & \textbf{0.30}  \\
\bottomrule
\end{tabular}
    } 
}
\label{tab:abl}
\end{table}


\section{Conclusion}
We introduced VLA-World, a unified Vision-Language-Action World Model that integrates predictive imagination and reflective reasoning for end-to-end autonomous driving. Unlike VLA or world models, VLA-World combines future simulation with reasoning, enabling the model to both anticipate and evaluate outcomes for safer decision-making.
To realize this framework, we constructed nuScenes-GR-20K, a dataset designed for future frame generation and reasoning, and developed a three-stage training pipeline including pretraining, supervised fine-tuning, and reinforcement learning with GRPO.
Extensive experiments demonstrate that VLA-World surpasses existing methods in trajectory planning and visual generation, achieving improved interpretability and human-like reasoning, paving the way toward more intelligent and reliable autonomous driving.

\newpage
\paragraph{Acknowledgements.} This work was supported in part by NSFC (62322113, 62376156), Shanghai Municipal Science and Technology Major Project (2025SHZDZX025G15, 2021SHZDZX0102), and the Fundamental Research Funds for the Central Universities.

{
    \small
    \bibliographystyle{ieeenat_fullname}
    \bibliography{main}

\begin{thebibliography}{87}
\providecommand{\natexlab}[1]{#1}
\providecommand{\url}[1]{\texttt{#1}}
\expandafter\ifx\csname urlstyle\endcsname\relax
  \providecommand{\doi}[1]{doi: #1}\else
  \providecommand{\doi}{doi: \begingroup \urlstyle{rm}\Url}\fi

\bibitem[Achiam et~al.(2023)Achiam, Adler, Agarwal, Ahmad, Akkaya, Aleman, Almeida, Altenschmidt, Altman, Anadkat, et~al.]{achiam2023gpt4}
Josh Achiam, Steven Adler, Sandhini Agarwal, Lama Ahmad, Ilge Akkaya, Florencia~Leoni Aleman, Diogo Almeida, Janko Altenschmidt, Sam Altman, Shyamal Anadkat, et~al.
\newblock Gpt-4 technical report.
\newblock \emph{arXiv preprint arXiv:2303.08774}, 2023.

\bibitem[Alayrac et~al.(2022)Alayrac, Donahue, Luc, Miech, Barr, Hasson, Lenc, Mensch, Millican, Reynolds, et~al.]{alayrac2022flamingo}
Jean-Baptiste Alayrac, Jeff Donahue, Pauline Luc, Antoine Miech, Iain Barr, Yana Hasson, Karel Lenc, Arthur Mensch, Katherine Millican, Malcolm Reynolds, et~al.
\newblock Flamingo: a visual language model for few-shot learning.
\newblock In \emph{NeurIPS}, pages 23716--23736, 2022.

\bibitem[Arai et~al.(2025)Arai, Miwa, Sasaki, Watanabe, Yamaguchi, Aoki, and Yamamoto]{arai2025covla}
Hidehisa Arai, Keita Miwa, Kento Sasaki, Kohei Watanabe, Yu Yamaguchi, Shunsuke Aoki, and Issei Yamamoto.
\newblock Covla: Comprehensive vision-language-action dataset for autonomous driving.
\newblock In \emph{WACV}, pages 1933--1943, 2025.

\bibitem[Bai et~al.(2023)Bai, Bai, Yang, Wang, Tan, Wang, Lin, Zhou, and Zhou]{bai2023qwenvl}
Jinze Bai, Shuai Bai, Shusheng Yang, Shijie Wang, Sinan Tan, Peng Wang, Junyang Lin, Chang Zhou, and Jingren Zhou.
\newblock Qwen-vl: A frontier large vision-language model with versatile abilities.
\newblock \emph{arXiv preprint arXiv:2308.12966}, 2023.

\bibitem[Bai et~al.(2025)Bai, Chen, Liu, Wang, Ge, Song, Dang, Wang, Wang, Tang, Zhong, Zhu, Yang, Li, Wan, Wang, Ding, Fu, Xu, Ye, Zhang, Xie, Cheng, Zhang, Yang, Xu, and Lin]{bai2025qwen25vl}
Shuai Bai, Keqin Chen, Xuejing Liu, Jialin Wang, Wenbin Ge, Sibo Song, Kai Dang, Peng Wang, Shijie Wang, Jun Tang, Humen Zhong, Yuanzhi Zhu, Mingkun Yang, Zhaohai Li, Jianqiang Wan, Pengfei Wang, Wei Ding, Zheren Fu, Yiheng Xu, Jiabo Ye, Xi Zhang, Tianbao Xie, Zesen Cheng, Hang Zhang, Zhibo Yang, Haiyang Xu, and Junyang Lin.
\newblock Qwen2.5-vl technical report.
\newblock \emph{arXiv preprint arXiv:2502.13923}, 2025.

\bibitem[Belkhale et~al.(2024)Belkhale, Ding, Xiao, Sermanet, Vuong, Tompson, Chebotar, Dwibedi, and Sadigh]{belkhale2024rt}
Suneel Belkhale, Tianli Ding, Ted Xiao, Pierre Sermanet, Quon Vuong, Jonathan Tompson, Yevgen Chebotar, Debidatta Dwibedi, and Dorsa Sadigh.
\newblock Rt-h: Action hierarchies using language.
\newblock \emph{arXiv preprint arXiv:2403.01823}, 2024.

\bibitem[Black et~al.(2024)Black, Brown, Driess, Esmail, Equi, Finn, Fusai, Groom, Hausman, Ichter, et~al.]{black2024pi_0}
Kevin Black, Noah Brown, Danny Driess, Adnan Esmail, Michael Equi, Chelsea Finn, Niccolo Fusai, Lachy Groom, Karol Hausman, Brian Ichter, et~al.
\newblock pi\_0: A vision-language-action flow model for general robot control.
\newblock \emph{arXiv preprint arXiv:2410.24164}, 2024.

\bibitem[Brohan et~al.(2022)Brohan, Brown, Carbajal, Chebotar, Dabis, Finn, Gopalakrishnan, Hausman, Herzog, Hsu, et~al.]{brohan2022rt}
Anthony Brohan, Noah Brown, Justice Carbajal, Yevgen Chebotar, Joseph Dabis, Chelsea Finn, Keerthana Gopalakrishnan, Karol Hausman, Alex Herzog, Jasmine Hsu, et~al.
\newblock Rt-1: Robotics transformer for real-world control at scale.
\newblock \emph{arXiv preprint arXiv:2212.06817}, 2022.

\bibitem[Brooks et~al.(2024)Brooks, Peebles, Holmes, DePue, Guo, Jing, Schnurr, Taylor, Luhman, Luhman, et~al.]{brooks2024video}
Tim Brooks, Bill Peebles, Connor Holmes, Will DePue, Yufei Guo, Li Jing, David Schnurr, Joe Taylor, Troy Luhman, Eric Luhman, et~al.
\newblock Video generation models as world simulators.
\newblock \emph{OpenAI Blog}, 1\penalty0 (8):\penalty0 1, 2024.

\bibitem[Caesar et~al.(2020)Caesar, Bankiti, Lang, Vora, Liong, Xu, Krishnan, Pan, Baldan, and Beijbom]{caesar2020nuscenes}
Holger Caesar, Varun Bankiti, Alex~H Lang, Sourabh Vora, Venice~Erin Liong, Qiang Xu, Anush Krishnan, Yu Pan, Giancarlo Baldan, and Oscar Beijbom.
\newblock nuscenes: A multimodal dataset for autonomous driving.
\newblock In \emph{CVPR}, pages 11621--11631, 2020.

\bibitem[Cen et~al.(2025)Cen, Yu, Yuan, Jiang, Huang, Guo, Li, Song, Luo, Wang, et~al.]{cen2025worldvla}
Jun Cen, Chaohui Yu, Hangjie Yuan, Yuming Jiang, Siteng Huang, Jiayan Guo, Xin Li, Yibing Song, Hao Luo, Fan Wang, et~al.
\newblock Worldvla: Towards autoregressive action world model.
\newblock \emph{arXiv preprint arXiv:2506.21539}, 2025.

\bibitem[Chen et~al.(2024{\natexlab{a}})Chen, Wu, Chitta, Jaeger, Geiger, and Li]{chen2024end}
Li Chen, Penghao Wu, Kashyap Chitta, Bernhard Jaeger, Andreas Geiger, and Hongyang Li.
\newblock End-to-end autonomous driving: Challenges and frontiers.
\newblock \emph{IEEE TPAMI}, 2024{\natexlab{a}}.

\bibitem[Chen et~al.(2024{\natexlab{b}})Chen, Jiang, Gao, Liao, Xu, Zhang, Huang, Liu, and Wang]{vad2}
Shaoyu Chen, Bo Jiang, Hao Gao, Bencheng Liao, Qing Xu, Qian Zhang, Chang Huang, Wenyu Liu, and Xinggang Wang.
\newblock Vadv2: End-to-end vectorized autonomous driving via probabilistic planning, 2024{\natexlab{b}}.

\bibitem[Chen et~al.(2025)Chen, Wu, Liu, Pan, Liu, Xie, Yu, and Ruan]{chen2025janus}
Xiaokang Chen, Zhiyu Wu, Xingchao Liu, Zizheng Pan, Wen Liu, Zhenda Xie, Xingkai Yu, and Chong Ruan.
\newblock Janus-pro: Unified multimodal understanding and generation with data and model scaling.
\newblock \emph{arXiv preprint arXiv:2501.17811}, 2025.

\bibitem[Chen et~al.(2024{\natexlab{c}})Chen, Ye, Xu, Cao, and Chen]{chen2024ppad}
Zhili Chen, Maosheng Ye, Shuangjie Xu, Tongyi Cao, and Qifeng Chen.
\newblock Ppad: Iterative interactions of prediction and planning for end-to-end autonomous driving.
\newblock In \emph{ECCV}, pages 239--256, 2024{\natexlab{c}}.

\bibitem[Chi et~al.(2025)Chi, Gao, Liu, Liu, Liu, Li, Yang, Yu, Wang, Li, et~al.]{chi2025impromptu}
Haohan Chi, Huan-ang Gao, Ziming Liu, Jianing Liu, Chenyu Liu, Jinwei Li, Kaisen Yang, Yangcheng Yu, Zeda Wang, Wenyi Li, et~al.
\newblock Impromptu vla: Open weights and open data for driving vision-language-action models.
\newblock \emph{arXiv preprint arXiv:2505.23757}, 2025.

\bibitem[Driess et~al.(2023)Driess, Xia, Sajjadi, Lynch, Chowdhery, Wahid, Tompson, Vuong, Yu, Huang, et~al.]{driess2023palm}
Danny Driess, Fei Xia, Mehdi~SM Sajjadi, Corey Lynch, Aakanksha Chowdhery, Ayzaan Wahid, Jonathan Tompson, Quan Vuong, Tianhe Yu, Wenlong Huang, et~al.
\newblock Palm-e: An embodied multimodal language model.
\newblock In \emph{ICML}, 2023.

\bibitem[Esser et~al.(2021)Esser, Rombach, and Ommer]{esser2021taming}
Patrick Esser, Robin Rombach, and Bjorn Ommer.
\newblock Taming transformers for high-resolution image synthesis.
\newblock In \emph{CVPR}, pages 12873--12883, 2021.

\bibitem[Guo et~al.(2025)Guo, Yang, Zhang, Song, Wang, Zhu, Xu, Zhang, Ma, Bi, et~al.]{guo2025deepseek}
Daya Guo, Dejian Yang, Haowei Zhang, Junxiao Song, Peiyi Wang, Qihao Zhu, Runxin Xu, Ruoyu Zhang, Shirong Ma, Xiao Bi, et~al.
\newblock Deepseek-r1: Incentivizing reasoning capability in llms via reinforcement learning.
\newblock \emph{arXiv preprint arXiv:2501.12948}, 2025.

\bibitem[Han et~al.(2025)Han, Guo, Xu, and Shen]{han2025dme}
Wencheng Han, Dongqian Guo, Cheng-Zhong Xu, and Jianbing Shen.
\newblock Dme-driver: Integrating human decision logic and 3d scene perception in autonomous driving.
\newblock In \emph{AAAI}, pages 3347--3355, 2025.

\bibitem[Hassan et~al.(2025)Hassan, Stapf, Rahimi, Rezende, Haghighi, Br{\"u}ggemann, Katircioglu, Zhang, Chen, Saha, et~al.]{hassan2025gem}
Mariam Hassan, Sebastian Stapf, Ahmad Rahimi, Pedro Rezende, Yasaman Haghighi, David Br{\"u}ggemann, Isinsu Katircioglu, Lin Zhang, Xiaoran Chen, Suman Saha, et~al.
\newblock Gem: A generalizable ego-vision multimodal world model for fine-grained ego-motion, object dynamics, and scene composition control.
\newblock In \emph{CVPR}, pages 22404--22415, 2025.

\bibitem[Hu et~al.(2023{\natexlab{a}})Hu, Russell, Yeo, Murez, Fedoseev, Kendall, Shotton, and Corrado]{Hu2023GAIA1AG}
Anthony Hu, Lloyd Russell, Hudson Yeo, Zak Murez, George Fedoseev, Alex Kendall, Jamie Shotton, and Gianluca Corrado.
\newblock Gaia-1: A generative world model for autonomous driving.
\newblock \emph{arXiv preprint arXiv:2309.17080}, 2023{\natexlab{a}}.

\bibitem[Hu et~al.(2022)Hu, Chen, Wu, Li, Yan, and Tao]{hu2022stp3}
Shengchao Hu, Li Chen, Penghao Wu, Hongyang Li, Junchi Yan, and Dacheng Tao.
\newblock St-p3: End-to-end vision-based autonomous driving via spatial-temporal feature learning.
\newblock In \emph{ECCV}, pages 533--549, 2022.

\bibitem[Hu et~al.(2024)Hu, Yin, Jia, Deng, Guo, Zhang, Long, and Tan]{hu2024drivingworld}
Xiaotao Hu, Wei Yin, Mingkai Jia, Junyuan Deng, Xiaoyang Guo, Qian Zhang, Xiaoxiao Long, and Ping Tan.
\newblock Drivingworld: Constructing world model for autonomous driving via video gpt.
\newblock \emph{arXiv preprint arXiv:2412.19505}, 2024.

\bibitem[Hu et~al.(2023{\natexlab{b}})Hu, Yang, Chen, Li, Sima, Zhu, Chai, Du, Lin, Wang, Lu, Jia, Liu, Dai, Qiao, and Li]{uniad}
Yihan Hu, Jiazhi Yang, Li Chen, Keyu Li, Chonghao Sima, Xizhou Zhu, Siqi Chai, Senyao Du, Tianwei Lin, Wenhai Wang, Lewei Lu, Xiaosong Jia, Qiang Liu, Jifeng Dai, Yu Qiao, and Hongyang Li.
\newblock Planning-oriented autonomous driving.
\newblock In \emph{CVPR}, pages 17853--17862, 2023{\natexlab{b}}.

\bibitem[Huang et~al.(2024)Huang, Tang, Chen, Lin, and Jie]{RDA-Driver}
Zhijian Huang, Tao Tang, Shaoxiang Chen, Sihao Lin, and Zequn et~al. Jie.
\newblock Making large language models better planners with reasoning-decision alignment.
\newblock In \emph{ECCV}, pages 73--90, 2024.

\bibitem[Hwang et~al.(2024)Hwang, Xu, Lin, Hung, Ji, Choi, Huang, He, Covington, Sapp, et~al.]{hwang2024emma}
Jyh-Jing Hwang, Runsheng Xu, Hubert Lin, Wei-Chih Hung, Jingwei Ji, Kristy Choi, Di Huang, Tong He, Paul Covington, Benjamin Sapp, et~al.
\newblock Emma: End-to-end multimodal model for autonomous driving.
\newblock \emph{arXiv preprint arXiv:2410.23262}, 2024.

\bibitem[Jia et~al.(2023)Jia, Mao, Liu, Zhao, Wen, Zhang, Zhang, and Wang]{jia2023adriver}
Fan Jia, Weixin Mao, Yingfei Liu, Yucheng Zhao, Yuqing Wen, Chi Zhang, Xiangyu Zhang, and Tiancai Wang.
\newblock Adriver-i: A general world model for autonomous driving.
\newblock \emph{arXiv:2311.13549}, 2023.

\bibitem[Jiang et~al.(2023)Jiang, Chen, Xu, Liao, Chen, Zhou, Zhang, Liu, Huang, and Wang]{vad}
Bo Jiang, Shaoyu Chen, Qing Xu, Bencheng Liao, Jiajie Chen, Helong Zhou, Qian Zhang, Wenyu Liu, Chang Huang, and Xinggang Wang.
\newblock Vad: Vectorized scene representation for efficient autonomous driving.
\newblock In \emph{ICCV}, pages 8306--8316, 2023.

\bibitem[Jiang et~al.(2025)Jiang, Chen, Zhang, Liu, and Wang]{alphadrive}
Bo Jiang, Shaoyu Chen, Qian Zhang, Wenyu Liu, and Xinggang Wang.
\newblock Alphadrive: Unleashing the power of vlms in autonomous driving via reinforcement learning and reasoning.
\newblock \emph{arXiv preprint arXiv:2503.07608}, 2025.

\bibitem[Kim et~al.(2021)Kim, Philion, Torralba, and Fidler]{kim2021drivegan}
Seung~Wook Kim, Jonah Philion, Antonio Torralba, and Sanja Fidler.
\newblock Drivegan: Towards a controllable high-quality neural simulation.
\newblock In \emph{CVPR}, pages 5820--5829, 2021.

\bibitem[Li et~al.(2024{\natexlab{a}})Li, Guo, Liu, Zou, Ding, Chen, Zhu, Tan, Zhang, Wang, et~al.]{li2024uniscene}
Bohan Li, Jiazhe Guo, Hongsi Liu, Yingshuang Zou, Yikang Ding, Xiwu Chen, Hu Zhu, Feiyang Tan, Chi Zhang, Tiancai Wang, et~al.
\newblock Uniscene: Unified occupancy-centric driving scene generation.
\newblock \emph{arXiv preprint arXiv:2412.05435}, 2024{\natexlab{a}}.

\bibitem[Li et~al.(2024{\natexlab{b}})Li, Wang, Mao, Ivanovic, Veer, Leung, and Pavone]{li2024llada}
Boyi Li, Yue Wang, Jiageng Mao, Boris Ivanovic, Sushant Veer, Karen Leung, and Marco Pavone.
\newblock Driving everywhere with large language model policy adaptation.
\newblock In \emph{CVPR}, pages 14948--14957, 2024{\natexlab{b}}.

\bibitem[Li et~al.(2025{\natexlab{a}})Li, Ma, Du, Peng, Liang, Liu, Ma, Jin, Zhao, Zeng, et~al.]{li2025omninwm}
Bohan Li, Zhuang Ma, Dalong Du, Baorui Peng, Zhujin Liang, Zhenqiang Liu, Chao Ma, Yueming Jin, Hao Zhao, Wenjun Zeng, et~al.
\newblock Omninwm: Omniscient driving navigation world models.
\newblock \emph{arXiv preprint arXiv:2510.18313}, 2025{\natexlab{a}}.

\bibitem[Li et~al.(2022)Li, Li, Xiong, and Hoi]{li2022blip}
Junnan Li, Dongxu Li, Caiming Xiong, and Steven Hoi.
\newblock Blip: Bootstrapping language-image pre-training for unified vision-language understanding and generation.
\newblock In \emph{ICML}, pages 12888--12900, 2022.

\bibitem[Li et~al.(2023)Li, Li, Savarese, and Hoi]{li2023blip2}
Junnan Li, Dongxu Li, Silvio Savarese, and Steven Hoi.
\newblock Blip-2: Bootstrapping language-image pre-training with frozen image encoders and large language models.
\newblock In \emph{ICML}, pages 19730--19742, 2023.

\bibitem[Li et~al.(2025{\natexlab{b}})Li, Li, Zheng, Sun, Wang, and Chen]{li2025semi}
Xiang Li, Pengfei Li, Yupeng Zheng, Wei Sun, Yan Wang, and Yilun Chen.
\newblock Semi-supervised vision-centric 3d occupancy world model for autonomous driving.
\newblock \emph{arXiv preprint arXiv:2502.07309}, 2025{\natexlab{b}}.

\bibitem[Li et~al.(2024{\natexlab{c}})Li, Yu, Lan, Li, Kautz, Lu, and Alvarez]{bev-planner}
Zhiqi Li, Zhiding Yu, Shiyi Lan, Jiahan Li, Jan Kautz, Tong Lu, and Jose~M. Alvarez.
\newblock Is ego status all you need for open-loop end-to-end autonomous driving?
\newblock In \emph{CVPR}, pages 14864--14873, 2024{\natexlab{c}}.

\bibitem[Liu et~al.(2023)Liu, Li, Wu, and Lee]{liu2024llava}
Haotian Liu, Chunyuan Li, Qingyang Wu, and Yong~Jae Lee.
\newblock Visual instruction tuning.
\newblock In \emph{NeurIPS}, pages 34892--34916, 2023.

\bibitem[Liu et~al.(2024)Liu, Li, Li, and Lee]{liu2024improved}
Haotian Liu, Chunyuan Li, Yuheng Li, and Yong~Jae Lee.
\newblock Improved baselines with visual instruction tuning.
\newblock In \emph{CVPR}, pages 26296--26306, 2024.

\bibitem[Lu et~al.(2024)Lu, Huang, Yang, Zhang, and Zhang]{lu2025wovogen}
Jiachen Lu, Ze Huang, Zeyu Yang, Jiahui Zhang, and Li Zhang.
\newblock Wovogen: World volume-aware diffusion for controllable multi-camera driving scene generation.
\newblock In \emph{ECCV}, pages 329--345, 2024.

\bibitem[Ma et~al.(2024)Ma, Cao, Sun, Pavone, and Xiao]{ma2024dolphins}
Yingzi Ma, Yulong Cao, Jiachen Sun, Marco Pavone, and Chaowei Xiao.
\newblock Dolphins: Multimodal language model for driving.
\newblock In \emph{ECCV}, pages 403--420, 2024.

\bibitem[Mao et~al.(2023)Mao, Ye, Qian, Pavone, and Wang]{mao2023language}
Jiageng Mao, Junjie Ye, Yuxi Qian, Marco Pavone, and Yue Wang.
\newblock A language agent for autonomous driving.
\newblock \emph{arXiv preprint arXiv:2311.10813}, 2023.

\bibitem[Min et~al.(2024)Min, Zhao, Xiao, Zhao, Xu, Zhu, Jin, Li, Guo, Xing, et~al.]{min2024driveworld}
Chen Min, Dawei Zhao, Liang Xiao, Jian Zhao, Xinli Xu, Zheng Zhu, Lei Jin, Jianshu Li, Yulan Guo, Junliang Xing, et~al.
\newblock Driveworld: 4d pre-trained scene understanding via world models for autonomous driving.
\newblock In \emph{CVPR}, pages 15522--15533, 2024.

\bibitem[Nie et~al.(2024)Nie, Peng, Wang, Cai, Han, Xu, and Zhang]{nie2024reason2drive}
Ming Nie, Renyuan Peng, Chunwei Wang, Xinyue Cai, Jianhua Han, Hang Xu, and Li Zhang.
\newblock Reason2drive: Towards interpretable and chain-based reasoning for autonomous driving.
\newblock In \emph{ECCV}, pages 292--308, 2024.

\bibitem[Qian et~al.(2025)Qian, Jiang, Zhong, Luo, Huang, Zhu, Jiang, Yang, Fu, Miao, et~al.]{qian2025agentthink}
Kangan Qian, Sicong Jiang, Yang Zhong, Ziang Luo, Zilin Huang, Tianze Zhu, Kun Jiang, Mengmeng Yang, Zheng Fu, Jinyu Miao, et~al.
\newblock Agentthink: A unified framework for tool-augmented chain-of-thought reasoning in vision-language models for autonomous driving.
\newblock \emph{arXiv preprint arXiv:2505.15298}, 2025.

\bibitem[Ren et~al.(2025)Ren, Wang, Hou, Tang, Wang, and Ma]{ren2025grounding}
Xiangxuan Ren, Zhongdao Wang, Liping Hou, Pin Tang, Guoqing Wang, and Chao Ma.
\newblock Grounding everything in tokens for multimodal large language models.
\newblock \emph{arXiv preprint arXiv:2512.10554}, 2025.

\bibitem[Sarkar et~al.(2025)Sarkar, Idris, and Yu]{sarkar2025reasoning}
Ayushman Sarkar, Mohd Yamani~Idna Idris, and Zhenyu Yu.
\newblock Reasoning in computer vision: Taxonomy, models, tasks, and methodologies.
\newblock \emph{arXiv preprint arXiv:2508.10523}, 2025.

\bibitem[Schulman et~al.(2017)Schulman, Wolski, Dhariwal, Radford, and Klimov]{schulman2017proximal}
John Schulman, Filip Wolski, Prafulla Dhariwal, Alec Radford, and Oleg Klimov.
\newblock Proximal policy optimization algorithms.
\newblock \emph{arXiv preprint arXiv:1707.06347}, 2017.

\bibitem[Shao et~al.(2024{\natexlab{a}})Shao, Hu, Wang, Song, Waslander, Liu, and Li]{shao2024lmdrive}
Hao Shao, Yuxuan Hu, Letian Wang, Guanglu Song, Steven~L Waslander, Yu Liu, and Hongsheng Li.
\newblock Lmdrive: Closed-loop end-to-end driving with large language models.
\newblock In \emph{CVPR}, pages 15120--15130, 2024{\natexlab{a}}.

\bibitem[Shao et~al.(2024{\natexlab{b}})Shao, Wang, Zhu, Xu, Song, Bi, Zhang, Zhang, Li, Wu, et~al.]{grpo}
Zhihong Shao, Peiyi Wang, Qihao Zhu, Runxin Xu, Junxiao Song, Xiao Bi, Haowei Zhang, Mingchuan Zhang, YK Li, Y Wu, et~al.
\newblock Deepseekmath: Pushing the limits of mathematical reasoning in open language models.
\newblock \emph{arXiv preprint arXiv:2402.03300}, 2024{\natexlab{b}}.

\bibitem[Sheng et~al.(2024)Sheng, Zhang, Ye, Wu, Zhang, Zhang, Peng, Lin, and Wu]{sheng2024hybridflow}
Guangming Sheng, Chi Zhang, Zilingfeng Ye, Xibin Wu, Wang Zhang, Ru Zhang, Yanghua Peng, Haibin Lin, and Chuan Wu.
\newblock Hybridflow: A flexible and efficient rlhf framework.
\newblock \emph{arXiv preprint arXiv: 2409.19256}, 2024.

\bibitem[Tian et~al.(2024)Tian, Gu, Li, Liu, Hu, Wang, Zhan, Jia, Lang, and Zhao]{tian2024drivevlm}
Xiaoyu Tian, Junru Gu, Bailin Li, Yicheng Liu, Chenxu Hu, Yang Wang, Kun Zhan, Peng Jia, Xianpeng Lang, and Hang Zhao.
\newblock Drivevlm: The convergence of autonomous driving and large vision-language models.
\newblock \emph{arXiv preprint arXiv:2402.12289}, 2024.

\bibitem[Touvron et~al.(2023)Touvron, Lavril, Izacard, Martinet, Lachaux, Lacroix, Rozi{\`e}re, Goyal, Hambro, Azhar, et~al.]{touvron2023llama}
Hugo Touvron, Thibaut Lavril, Gautier Izacard, Xavier Martinet, Marie-Anne Lachaux, Timoth{\'e}e Lacroix, Baptiste Rozi{\`e}re, Naman Goyal, Eric Hambro, Faisal Azhar, et~al.
\newblock Llama: Open and efficient foundation language models.
\newblock \emph{arXiv preprint arXiv:2302.13971}, 2023.

\bibitem[Van Den~Oord et~al.(2017)Van Den~Oord, Vinyals, et~al.]{van2017neural}
Aaron Van Den~Oord, Oriol Vinyals, et~al.
\newblock Neural discrete representation learning.
\newblock In \emph{NeurIPS}, pages 6309--6318, 2017.

\bibitem[Wang et~al.(2024{\natexlab{a}})Wang, Bai, Tan, Wang, Fan, Bai, Chen, Liu, Wang, Ge, et~al.]{wang2024qwen2vl}
Peng Wang, Shuai Bai, Sinan Tan, Shijie Wang, Zhihao Fan, Jinze Bai, Keqin Chen, Xuejing Liu, Jialin Wang, Wenbin Ge, et~al.
\newblock Qwen2-vl: Enhancing vision-language model's perception of the world at any resolution.
\newblock \emph{arXiv preprint arXiv:2409.12191}, 2024{\natexlab{a}}.

\bibitem[Wang et~al.(2025)Wang, Yu, Jiang, Lan, Shi, Chang, Kautz, Li, and Alvarez]{wang2025omnidrive}
Shihao Wang, Zhiding Yu, Xiaohui Jiang, Shiyi Lan, Min Shi, Nadine Chang, Jan Kautz, Ying Li, and Jose~M Alvarez.
\newblock Omnidrive: A holistic vision-language dataset for autonomous driving with counterfactual reasoning.
\newblock In \emph{CVPR}, pages 22442--22452, 2025.

\bibitem[Wang et~al.(2023{\natexlab{a}})Wang, Xie, Hu, Zou, Fan, Tong, Wen, Wu, Deng, Li, et~al.]{drivemlm}
Wenhai Wang, Jiangwei Xie, ChuanYang Hu, Haoming Zou, Jianan Fan, Wenwen Tong, Yang Wen, Silei Wu, Hanming Deng, Zhiqi Li, et~al.
\newblock Drivemlm: Aligning multi-modal large language models with behavioral planning states for autonomous driving.
\newblock \emph{arXiv preprint arXiv:2312.09245}, 2023{\natexlab{a}}.

\bibitem[Wang et~al.(2023{\natexlab{b}})Wang, Zhu, Huang, Chen, Zhu, and Lu]{wang2023drivedreamer}
Xiaofeng Wang, Zheng Zhu, Guan Huang, Xinze Chen, Jiagang Zhu, and Jiwen Lu.
\newblock Drivedreamer: Towards real-world-driven world models for autonomous driving.
\newblock \emph{arXiv:2309.09777}, 2023{\natexlab{b}}.

\bibitem[Wang et~al.(2024{\natexlab{b}})Wang, Zhu, Huang, Wang, Chen, and Lu]{wang2024worlddreamer}
Xiaofeng Wang, Zheng Zhu, Guan Huang, Boyuan Wang, Xinze Chen, and Jiwen Lu.
\newblock Worlddreamer: Towards general world models for video generation via predicting masked tokens.
\newblock \emph{arXiv:2401.09985}, 2024{\natexlab{b}}.

\bibitem[Wang et~al.(2024{\natexlab{c}})Wang, He, Fan, Li, Chen, and Zhang]{wang2024driving}
Yuqi Wang, Jiawei He, Lue Fan, Hongxin Li, Yuntao Chen, and Zhaoxiang Zhang.
\newblock Driving into the future: Multiview visual forecasting and planning with world model for autonomous driving.
\newblock In \emph{CVPR}, pages 14749--14759, 2024{\natexlab{c}}.

\bibitem[Wei et~al.(2022)Wei, Wang, Schuurmans, Bosma, Chi, Xia, Le, and Zhou]{wei2022chain}
Jason Wei, Xuezhi Wang, Dale Schuurmans, Maarten Bosma, Ed~H. Chi, F. Xia, Quoc Le, and Denny Zhou.
\newblock Chain of thought prompting elicits reasoning in large language models.
\newblock In \emph{NeurIPS}, pages 24824--24837, 2022.

\bibitem[Wu et~al.(2025)Wu, Chen, Wu, Ma, Liu, Pan, Liu, Xie, Yu, Ruan, et~al.]{wu2025janus}
Chengyue Wu, Xiaokang Chen, Zhiyu Wu, Yiyang Ma, Xingchao Liu, Zizheng Pan, Wen Liu, Zhenda Xie, Xingkai Yu, Chong Ruan, et~al.
\newblock Janus: Decoupling visual encoding for unified multimodal understanding and generation.
\newblock In \emph{CVPR}, pages 12966--12977, 2025.

\bibitem[Wu et~al.(2023)Wu, Liu, Luo, Zhong, Chen, Xiao, Hou, Lou, Chen, Yang, et~al.]{wu2023mars}
Zirui Wu, Tianyu Liu, Liyi Luo, Zhide Zhong, Jianteng Chen, Hongmin Xiao, Chao Hou, Haozhe Lou, Yuantao Chen, Runyi Yang, et~al.
\newblock Mars: An instance-aware, modular and realistic simulator for autonomous driving.
\newblock In \emph{CAAI}, pages 3--15, 2023.

\bibitem[Xie et~al.(2024)Xie, Mao, Bai, Zhang, Wang, Lin, Gu, Chen, Yang, and Shou]{xie2024show}
Jinheng Xie, Weijia Mao, Zechen Bai, David~Junhao Zhang, Weihao Wang, Kevin~Qinghong Lin, Yuchao Gu, Zhijie Chen, Zhenheng Yang, and Mike~Zheng Shou.
\newblock Show-o: One single transformer to unify multimodal understanding and generation.
\newblock \emph{arXiv preprint arXiv:2408.12528}, 2024.

\bibitem[Xu et~al.(2025)Xu, Lu, Yan, Cai, Liu, and Chen]{xu2025occ}
Tianshuo Xu, Hao Lu, Xu Yan, Yingjie Cai, Bingbing Liu, and Yingcong Chen.
\newblock Occ-llm: Enhancing autonomous driving with occupancy-based large language models.
\newblock \emph{arXiv:2502.06419}, 2025.

\bibitem[Xu et~al.(2023)Xu, Zhang, Xie, Zhao, Guo, Wong, Li, and Zhao]{drivegpt4}
Zhenhua Xu, Yujia Zhang, Enze Xie, Zhen Zhao, Yong Guo, Kwan-Yee~K Wong, Zhenguo Li, and Hengshuang Zhao.
\newblock Drivegpt4: Interpretable end-to-end autonomous driving via large language model.
\newblock \emph{arXiv preprint arXiv:2310.01412}, 2023.

\bibitem[Yang et~al.(2024{\natexlab{a}})Yang, Yang, Hui, Zheng, Yu, Zhou, Li, Li, Liu, Huang, et~al.]{yang2024qwen2}
An Yang, Baosong Yang, Binyuan Hui, Bo Zheng, Bowen Yu, Chang Zhou, Chengpeng Li, Chengyuan Li, Dayiheng Liu, Fei Huang, et~al.
\newblock Qwen2 technical report.
\newblock \emph{arXiv preprint arXiv:2407.10671}, 2024{\natexlab{a}}.

\bibitem[Yang et~al.(2024{\natexlab{b}})Yang, Gao, Qiu, Chen, Li, Dai, Chitta, Wu, Zeng, Luo, et~al.]{yang2024generalized}
Jiazhi Yang, Shenyuan Gao, Yihang Qiu, Li Chen, Tianyu Li, Bo Dai, Kashyap Chitta, Penghao Wu, Jia Zeng, Ping Luo, et~al.
\newblock Generalized predictive model for autonomous driving.
\newblock In \emph{CVPR}, pages 14662--14672, 2024{\natexlab{b}}.

\bibitem[Yang et~al.(2025{\natexlab{a}})Yang, Chitta, Gao, Chen, Shao, Jia, Li, Geiger, Yue, and Chen]{yang2025resim}
Jiazhi Yang, Kashyap Chitta, Shenyuan Gao, Long Chen, Yuqian Shao, Xiaosong Jia, Hongyang Li, Andreas Geiger, Xiangyu Yue, and Li Chen.
\newblock Resim: Reliable world simulation for autonomous driving.
\newblock \emph{arXiv preprint arXiv:2506.09981}, 2025{\natexlab{a}}.

\bibitem[Yang et~al.(2025{\natexlab{b}})Yang, Mei, Ma, Du, Chen, Qian, Feng, and Liu]{yang2025driving}
Yu Yang, Jianbiao Mei, Yukai Ma, Siliang Du, Wenqing Chen, Yijie Qian, Yuxiang Feng, and Yong Liu.
\newblock Driving in the occupancy world: Vision-centric 4d occupancy forecasting and planning via world models for autonomous driving.
\newblock In \emph{AAAI}, pages 9327--9335, 2025{\natexlab{b}}.

\bibitem[Yang et~al.(2025{\natexlab{c}})Yang, Chai, Jia, Li, Shao, Zhu, Su, and Yan]{drivemoe}
Zhenjie Yang, Yilin Chai, Xiaosong Jia, Qifeng Li, Yuqian Shao, Xuekai Zhu, Haisheng Su, and Junchi Yan.
\newblock Drivemoe: Mixture-of-experts for vision-language-action model in end-to-end autonomous driving.
\newblock \emph{arXiv preprint arXiv:2505.16278}, 2025{\natexlab{c}}.

\bibitem[Yuan et~al.(2025)Yuan, Tang, Luo, Chen, Qian, Sun, Chu, Cai, Zhang, and Li]{yuan2025autodrive}
Zhenlong Yuan, Jing Tang, Jinguo Luo, Rui Chen, Chengxuan Qian, Lei Sun, Xiangxiang Chu, Yujun Cai, Dapeng Zhang, and Shuo Li.
\newblock Autodrive-r$^2$: Incentivizing reasoning and self-reflection capacity for vla model in autonomous driving.
\newblock \emph{arXiv preprint arXiv:2509.01944}, 2025.

\bibitem[Zeng et~al.(2025)Zeng, Chang, Xie, Liu, Bai, Pan, Xu, and Wei]{zeng2025futuresightdrive}
Shuang Zeng, Xinyuan Chang, Mengwei Xie, Xinran Liu, Yifan Bai, Zheng Pan, Mu Xu, and Xing Wei.
\newblock Futuresightdrive: Thinking visually with spatio-temporal cot for autonomous driving.
\newblock \emph{arXiv preprint arXiv:2505.17685}, 2025.

\bibitem[Zhang et~al.(2024{\natexlab{a}})Zhang, Huang, Ray, and Ohn-Bar]{zhang2023coaching}
Jimuyang Zhang, Zanming Huang, Arijit Ray, and Eshed Ohn-Bar.
\newblock Feedback-guided autonomous driving.
\newblock In \emph{CVPR}, pages 15000--15011, 2024{\natexlab{a}}.

\bibitem[Zhang et~al.(2024{\natexlab{b}})Zhang, Xu, and Li]{zhang2024chatscene}
Jiawei Zhang, Chejian Xu, and Bo Li.
\newblock Chatscene: Knowledge-enabled safety-critical scenario generation for autonomous vehicles.
\newblock In \emph{CVPR}, pages 15459--15469, 2024{\natexlab{b}}.

\bibitem[Zhang et~al.(2025)Zhang, Liu, Qi, Wang, Yu, Zhang, Dong, He, Lu, Wang, et~al.]{zhang2025dreamvla}
Wenyao Zhang, Hongsi Liu, Zekun Qi, Yunnan Wang, Xinqiang Yu, Jiazhao Zhang, Runpei Dong, Jiawei He, Fan Lu, He Wang, et~al.
\newblock Dreamvla: a vision-language-action model dreamed with comprehensive world knowledge.
\newblock \emph{arXiv preprint arXiv:2507.04447}, 2025.

\bibitem[Zheng et~al.(2024{\natexlab{a}})Zheng, Chen, Huang, Zhang, Duan, and Lu]{zheng2024occworld}
Wenzhao Zheng, Weiliang Chen, Yuanhui Huang, Borui Zhang, Yueqi Duan, and Jiwen Lu.
\newblock Occworld: Learning a 3d occupancy world model for autonomous driving.
\newblock In \emph{ECCV}, pages 55--72, 2024{\natexlab{a}}.

\bibitem[Zheng et~al.(2024{\natexlab{b}})Zheng, Song, Guo, Zhang, and Chen]{zheng2024genad}
Wenzhao Zheng, Ruiqi Song, Xianda Guo, Chenming Zhang, and Long Chen.
\newblock Genad: Generative end-to-end autonomous driving.
\newblock \emph{arXiv preprint arXiv:2402.11502}, 2024{\natexlab{b}}.

\bibitem[Zheng et~al.(2024{\natexlab{c}})Zheng, Xia, Huang, Zuo, Zhou, and Lu]{zheng2024doe}
Wenzhao Zheng, Zetian Xia, Yuanhui Huang, Sicheng Zuo, Jie Zhou, and Jiwen Lu.
\newblock Doe-1: Closed-loop autonomous driving with large world model.
\newblock \emph{arXiv preprint arXiv:2412.09627}, 2024{\natexlab{c}}.

\bibitem[Zheng et~al.(2025{\natexlab{a}})Zheng, Mao, Ye, Li, Zhan, Lang, and Zhao]{zheng2025driveagent}
Weicheng Zheng, Xiaofei Mao, Nanfei Ye, Pengxiang Li, Kun Zhan, Xianpeng Lang, and Hang Zhao.
\newblock Driveagent-r1: Advancing vlm-based autonomous driving with hybrid thinking and active perception.
\newblock \emph{arXiv preprint arXiv:2507.20879}, 2025{\natexlab{a}}.

\bibitem[Zheng et~al.(2024{\natexlab{d}})Zheng, Zhang, Zhang, Ye, and Luo]{zheng2024llamafactory}
Yaowei Zheng, Richong Zhang, Junhao Zhang, Yanhan Ye, and Zheyan Luo.
\newblock Llamafactory: Unified efficient fine-tuning of 100+ language models.
\newblock In \emph{ACL}, pages 400--410, 2024{\natexlab{d}}.

\bibitem[Zheng et~al.(2025{\natexlab{b}})Zheng, Yang, Xing, Zhang, Zheng, Gao, Li, Zhang, Xia, Jia, et~al.]{zheng2025world4drive}
Yupeng Zheng, Pengxuan Yang, Zebin Xing, Qichao Zhang, Yuhang Zheng, Yinfeng Gao, Pengfei Li, Teng Zhang, Zhongpu Xia, Peng Jia, et~al.
\newblock World4drive: End-to-end autonomous driving via intention-aware physical latent world model.
\newblock In \emph{ICCV}, pages 28632--28642, 2025{\natexlab{b}}.

\bibitem[Zhou et~al.(2025{\natexlab{a}})Zhou, Han, Yang, Ma, and Knoll]{zhou2025opendrivevla}
Xingcheng Zhou, Xuyuan Han, Feng Yang, Yunpu Ma, and Alois~C Knoll.
\newblock Opendrivevla: Towards end-to-end autonomous driving with large vision language action model.
\newblock \emph{arXiv preprint arXiv:2503.23463}, 2025{\natexlab{a}}.

\bibitem[Zhou et~al.(2024)Zhou, Huang, Bu, Zeng, Li, Qiu, Zhu, Guo, Qiao, and Li]{zhou2024embodied}
Yunsong Zhou, Linyan Huang, Qingwen Bu, Jia Zeng, Tianyu Li, Hang Qiu, Hongzi Zhu, Minyi Guo, Yu Qiao, and Hongyang Li.
\newblock Embodied understanding of driving scenarios.
\newblock In \emph{ECCV}, pages 129--148, 2024.

\bibitem[Zhou et~al.(2025{\natexlab{b}})Zhou, Cai, Zhao, Zhang, Huang, Zhou, and Ma]{zhou2025autovla}
Zewei Zhou, Tianhui Cai, Seth~Z Zhao, Yun Zhang, Zhiyu Huang, Bolei Zhou, and Jiaqi Ma.
\newblock Autovla: A vision-language-action model for end-to-end autonomous driving with adaptive reasoning and reinforcement fine-tuning.
\newblock \emph{arXiv preprint arXiv:2506.13757}, 2025{\natexlab{b}}.

\bibitem[Zitkovich et~al.(2023)Zitkovich, Yu, Xu, Xu, Xiao, Xia, Wu, Wohlhart, Welker, Wahid, et~al.]{zitkovich2023rt}
Brianna Zitkovich, Tianhe Yu, Sichun Xu, Peng Xu, Ted Xiao, Fei Xia, Jialin Wu, Paul Wohlhart, Stefan Welker, Ayzaan Wahid, et~al.
\newblock Rt-2: Vision-language-action models transfer web knowledge to robotic control.
\newblock In \emph{CoRL}, pages 2165--2183, 2023.

\end{thebibliography}
}
\appendix
\clearpage
\setcounter{page}{1}
\maketitlesupplementary

\noindent In the supplementary material, we first present the methodology detail of our proposed VLA-World, including Group Relative Policy Optimization (GRPO), short-term trajectory prediction, and theoretical analysis of VLA-World. Then, we provide the details of the datasets and implementation. Furthermore, we present additional experimental results to demonstrate the effectiveness of VLA-World.


\section{Method Details}
\label{sec:method_supp}

\subsection{Group Relative Policy Optimization}
In the final training stage of VLA-World, we adopt Group Relative Policy Optimization (GRPO)~\cite{grpo} to unleash the latent reasoning and decision-making capabilities. Unlike traditional PPO~\cite{schulman2017proximal}, which relies on a computationally heavy value function (Critic) that often struggles with high-dimensional visual dynamics, GRPO operates in a value-free paradigm. It leverages group-wise statistics to estimate baselines, significantly reducing memory overhead while stabilizing training.

For each driving scenario prompt $o$, the current policy $\pi_\theta$ samples a group of $G$ candidate rollouts (outputs), denoted as $\{o, o_1, \dots, o_G\}$. These candidates represent diverse reasoning paths, ranging from conservative yielding to assertive maneuvering. Instead of relying on a neural reward model, we employ a set of lightweight, rule-based verifiers to compute rewards. These include outcome rewards (e.g., collision checking, generation quality, temporal consistency) and format rewards (e.g., strict compliance with the required output structure). Each rollout is evaluated to produce a scalar reward set $\{r_1, r_2, \dots, r_G\}$.

To determine the relative quality of each reasoning path, we compute the normalized advantage for the $i$-th rollout within the group:
\begin{equation}
    A_i = \frac{r_i - \mu}{\sigma}, \quad \mu = \frac{1}{G} \sum_j r_j, \ \sigma = \mathrm{std}(r_1, \ldots, r_G)
\end{equation}
This group-based normalization effectively serves as a dynamic baseline, encouraging the model to prioritize trajectories that outperform the group average. The policy is then updated by maximizing the following surrogate objective:
\begin{equation}
    \begin{split}
        J(\theta) &= \mathbb{E}\left[ \frac{1}{G} \sum_{i=1}^G \min\left( \frac{\pi_\theta(\tau_i \mid o)}{\pi_{\theta_{\text{old}}}(\tau_i \mid o)} A_i, \text{clip} \right) \right] \\
        &\quad - \beta D_{\text{KL}}(\pi_\theta, \pi_{\text{old}}).
    \end{split}
\end{equation}
where the KL-divergence term ensures the policy does not deviate excessively from the reference model (the SFT checkpoint), preventing reward hacking.

By optimizing this objective, VLA-World effectively performs \textit{Self-Verification}: it learns to implicitly discard hallucinatory or unsafe trajectories and reinforces the internal \textit{chain-of-thought} that leads to compliant and safe driving behaviors. This mechanism allows the model to refine its logical consistency purely through rule-based feedback, resulting in a robust planner that is both explainable and physically grounded.

\subsection{Short-term Trajectory Prediction}
To ensure the synthesized future views are physically plausible and consistent with the vehicle's movement, we employ a physics-grounded trajectory predictor. This module estimates the ego-vehicle's future position $\hat{\mathbf{P}}_{t+\tau}$ at a look-ahead horizon $\tau$ (e.g., 0.5s), conditioned on both the historical state sequence $\mathcal{H}$ and the high-level mission goal $g$ (e.g., \textit{Left}). We formulate this prediction as a superposition of inertial dynamics and intentional control.

\noindent \textbf{Kinematic State Estimation.}
First, we extract the vehicle's instantaneous kinematic state from the discrete historical trajectory $\mathcal{H} = \{\mathbf{P}_{t-N}, \dots, \mathbf{P}_t\}$, where $\mathbf{P}_i \in \mathbb{R}^2$ denotes the coordinates in the ego-frame. We approximate the current velocity $\mathbf{v}_t$ and the historical inertial acceleration $\mathbf{a}_{\text{hist}}$ using a finite difference method:
\begin{equation}
    \mathbf{v}_t = \frac{\mathbf{P}_t - \mathbf{P}_{t-1}}{\Delta t}, \quad \mathbf{a}_{\text{hist}} = \frac{\mathbf{v}_t - \mathbf{v}_{t-1}}{\Delta t}
    \label{eq:state_estimation}
\end{equation}
where $\Delta t$ represents the sampling interval. The term $\mathbf{a}_{\text{hist}}$ captures the vehicle's momentum prior to any new control inputs.

\noindent \textbf{Intention-Driven Refinement.}
A pure constant-acceleration model often fails to capture sudden maneuvers dictated by the mission goal. To address this, we introduce a goal-conditioned acceleration term $\mathbf{a}_{\text{goal}}$. The navigational command $c$ is mapped to a target spatial offset or a virtual waypoint, implying a required trajectory deviation. We derive $\mathbf{a}_{\text{goal}}$ as the constant acceleration required to shift the vehicle from its current state $\mathcal{S}_t = \{\mathbf{P}_t, \mathbf{v}_t\}$ to the target state determined by $c$ within the horizon $\tau$.

\noindent \textbf{Fusion and Prediction.}
The final predicted trajectory is modeled as a linear fusion of the historical inertia and the future intention. We define the effective acceleration $\mathbf{a}_{\text{eff}}$ using an adaptive weighting factor $\lambda \in [0, 1]$:
\begin{equation}
    \mathbf{a}_{\text{eff}} = (1 - \lambda)\mathbf{a}_{\text{hist}} + \lambda \mathbf{a}_{\text{goal}}
    \label{eq:acceleration_fusion}
\end{equation}
where $\mathbf{a}_{\text{goal}} = \frac{2}{\tau^2}(\Delta \mathbf{P}_{\text{ideal}} - \mathbf{v}_t \tau)$, and $\Delta \mathbf{P}_{\text{ideal}}$ represents the theoretical displacement required by the command $c$. Consequently, the predicted position $\hat{\mathbf{P}}_{t+\tau}$ is computed via the kinematic equation:
\begin{equation}
    \hat{\mathbf{P}}_{t+\tau} = \mathbf{P}_t + \mathbf{v}_t \tau + \frac{1}{2} \mathbf{a}_{\text{eff}} \tau^2
    \label{eq:final_prediction}
\end{equation}
This formulation allows our model to seamlessly transition between momentum-based continuity (e.g., straight-line driving) and intention-based maneuvering (e.g., sharp turns), providing a robust geometric prior for the subsequent frame generation process.

\subsection{Theoretical Analysis of VLA-World}
\label{sec:method_theory}

In this section, we provide a theoretical analysis for VLA-World by formalizing autonomous driving as a joint optimization problem. We demonstrate that VLA-World aligns better with the driving objective than independent VLA or World Model paradigms.

\noindent\textbf{The Joint Modeling Objective.} The central object of autonomous driving is the joint distribution of the planned ego-trajectory $\tau_{t:t+H}$ and the anticipated short-term future environment $x_{t+1}$, conditioned on observation history $o_{1:t}$ and goal $g$.
According to the probability chain rule, this joint distribution factorizes as:
\begin{equation}
    \begin{split}
        p(\tau_{t:t+H}, x_{t+1}& \mid o_{1:t}, g) = \\
        &\underbrace{p(\tau_{t:t+H} \mid o_{1:t}, g)}_{\text{Policy (Decision)}} \cdot \underbrace{p(x_{t+1} \mid o_{1:t}, \tau_{t+1})}_{\text{World Model (Imagination)}}
    \end{split}
    \label{eq_joint}
\end{equation}
We define the ultimate driving objective $J(\omega)$ as the expected task return $R$ (aggregating safety, comfort, and rule compliance) over this joint distribution:
\begin{equation}
    \begin{split}
        J(\omega) = \mathbb{E}_{p_\omega(\tau, x \mid o, g)} \big[ R(\tau_{t:t+H}, x_{t+1}) \big]
    \end{split}
    \label{eq_joint_exp}
\end{equation}
Learning to drive is thus equivalent to learning a parameter set $\omega$ that shapes this joint distribution to maximize reward. VLA-World explicitly parameterizes and optimizes both factors in Eq.~\eqref{eq_joint}, whereas previous paradigms only address one.

\noindent\textbf{Analysis of VLA Models.} A pure VLA model implicitly integrates out the future state $x_{t+1}$, modeling only the marginal policy distribution:
\begin{equation}
    \pi_{\text{VLA}}(\tau \mid o, g) \approx \int p^\star(\tau, x \mid o, g) \, dx
\end{equation}
From a variational inference perspective, ignoring the explicit future state $x$ leads to a loose approximation of the optimal policy. For any auxiliary distribution $q(x \mid o, \tau)$ describing the environment dynamics, the log-likelihood of the optimal policy is bounded by the Evidence Lower Bound (ELBO):
\begin{equation}
    \log p^\star(\tau \mid o, g) \ge \mathbb{E}_{x \sim q} \big[ \log p^\star(\tau, x \mid o, g) - \log q(x \mid o, \tau) \big]
\end{equation}
Theoretical Insight: A VLA model that discards $x_{t+1}$ is mathematically equivalent to optimizing a loose lower bound where the predictive information about scene evolution is lost. It tries to match the marginal directly without understanding the underlying causal variable $x$.
In contrast, VLA-World models the joint numerator $p(\tau, x \mid o, g)$ directly. By explicitly generating $x_{t+1}$, VLA-World tightens this bound, effectively using the "imagined" future to reduce the uncertainty in policy estimation.

\noindent\textbf{Analysis of World Models.}
Classical world models focus on learning the transition dynamics $p_{\text{WM}}(x_{t+1} \mid o, \tau)$ via a reconstruction objective:
\begin{equation}
    J_{\text{WM}}(\theta) = \mathbb{E} \big[ -\log p_\theta(x_{t+1} \mid o, \tau) \big]
\end{equation}
Crucially, this objective is weakly coupled to the driving decision. A world model seeks to maximize pixel fidelity, not driving safety. The planning is typically performed by a separate search procedure on top of this frozen model:
\begin{equation}
    \tau^{\text{WM}} = \arg\max_\tau \mathbb{E}_{x \sim p_{\text{WM}}} [ R(\tau, x) ]
    \label{eq_wm_plan}
\end{equation}
Theoretical Insight: Any mismatch between generative accuracy (reconstruction) and planning utility (safety) creates a performance bottleneck. A high-fidelity simulation of a collision is valid for Eq.~\eqref{eq_wm_plan} but disastrous for the agent. Unlike VLA-World, pure world models do not back-propagate the decision reward $R$ into the model parameters $\theta$, leaving the \textit{imagination} disconnected from the \textit{consequence}.

\noindent\textbf{Analysis of VLA-World.}
VLA-World unifies the policy $\pi_\omega$ and world model $p_\omega$ into a single autoregressive transformer. The gradient of our objective $J(\omega)$ (from Eq.~\eqref{eq_joint_exp}) naturally decomposes to update both components:
\begin{equation}
    \begin{split}
        \nabla_\omega &J(\omega) = \\
        &\mathbb{E} \Big[ \underbrace{\nabla_\omega \log \pi_\omega(\tau \mid o, g) \cdot R}_{\text{Policy Gradient}} + \underbrace{\nabla_\omega \log p_\omega(x \mid o, \tau) \cdot R}_{\text{World Model Gradient}} \Big]
    \end{split}
\end{equation}
This reveals the core mechanism: Both the decision term and the imagination term are optimized by the same driving reward $R$.
In our implementation, we employ reinforcement learning with GRPO. Let $u$ denote the full sequence of tokens including trajectory $\tau$, future frame $x$, and reasoning language. We maximize:
\begin{equation}
    \mathcal{J}_{\text{GRPO}}(\omega) = \mathbb{E}_{u \sim \pi_\omega} \big[ \log \pi_\omega(u \mid o, g) \cdot A(u) \big]
\end{equation}
Because $u$ contains the future-image tokens, the \textit{imagination} is no longer just minimizing reconstruction error; it is being reinforced to generate futures that lead to high-reward outcomes (e.g., highlighting risks that aid safety). This forms an imagination-decision loop.
Finally, we show that VLA-World is a strictly more expressive hypothesis class than either baseline.

\noindent\textbf{VLA as a special case:} If we mask the imagination branch (force $p_\omega(x \mid \dots)$ to be a delta function or ignore it), Eq.~\eqref{eq_joint} collapses to the marginal policy $\pi_\omega(\tau \mid o, g)$, recovering a standard VLA.

\noindent\textbf{World Model as a special case:} If we freeze the parameters of $p_\omega(x \mid o, \tau)$ and use an external optimizer for $\tau$, we recover the trajectory search of standard World Models (Eq.~\eqref{eq_wm_plan}).

\section{Experiments}
\subsection{Dataset}
We evaluate trajectory planning and future frame generation on the nuScenes dataset~\cite{caesar2020nuscenes} following the traditional end-to-end methods~\cite{uniad, vad, vad2}, VLA~\cite{wang2025omnidrive, zeng2025futuresightdrive, hwang2024emma} and world models~\cite{wang2023drivedreamer,kim2021drivegan,wang2024driving, zheng2024doe}. The nuScenes comprises 1,000 driving scenes, each about 20 seconds long, recorded with a 32-beam LiDAR and six cameras offering a full 360-degree view. The dataset includes 28,130 training samples, 6,019 validation samples, and 193,082 unlabeled samples. 

\begin{figure}[!t]
    \centering
    \captionsetup{type=figure}
    \includegraphics[width=1.0\linewidth]{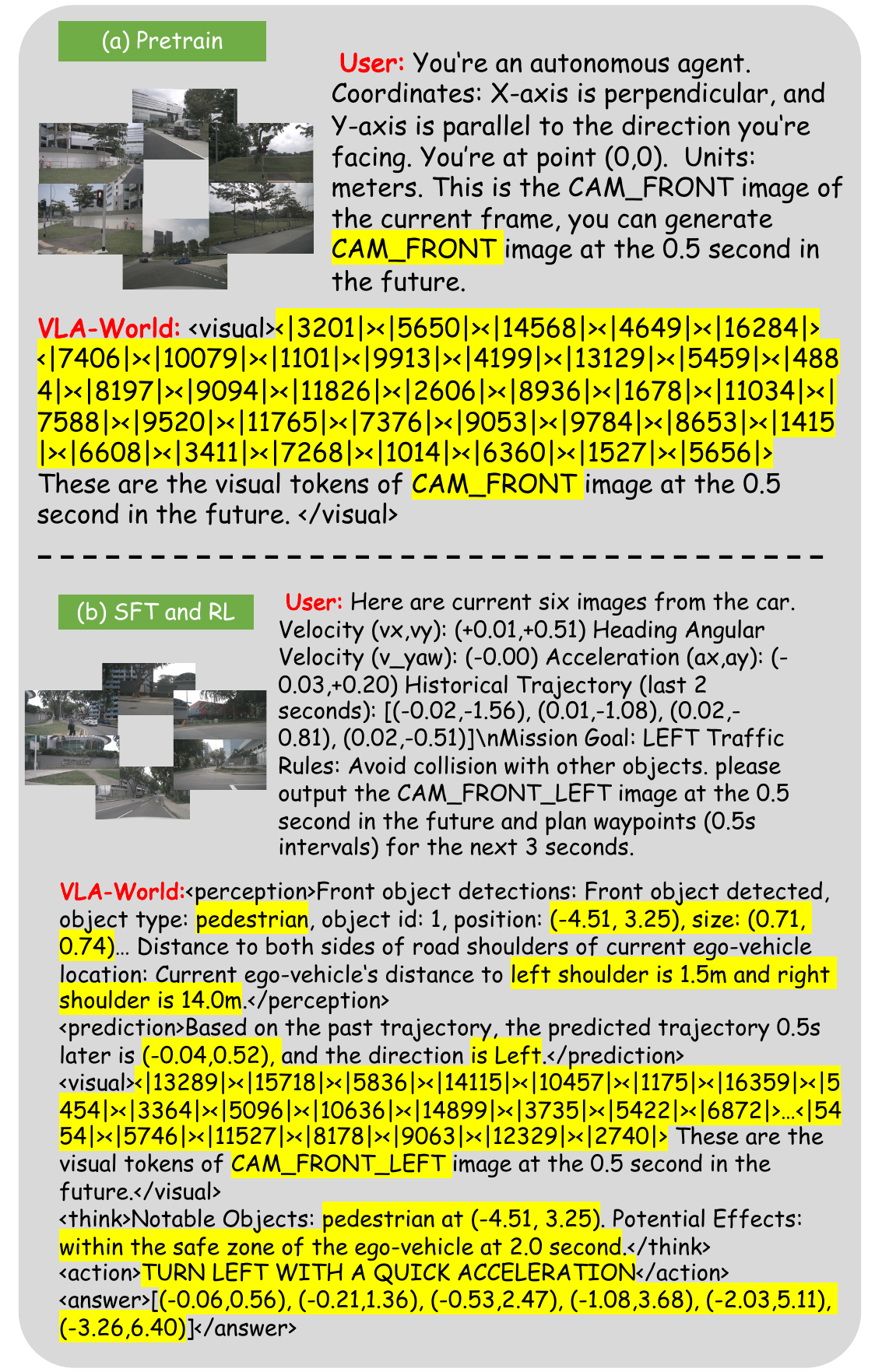}
    \caption{Data sample of (a) pretraining stage, (b) supervised fine-tuning, and (c) reinforcement learning. }
    \label{fig: supp_data}
\end{figure}

\noindent\textbf{Pretraining Stage.}
To endow the VLM with an intuitive understanding of physical dynamics, we construct a visual generation pretraining dataset as shown in Fig.~\ref{fig: supp_data} (a) ($\approx$500k). In this stage, the model functions strictly as a generative world model. The input prompt consists of current multi-view observations alongside explicit definitions of the ego-centric coordinate system and physical units. The objective is to autoregressively predict the discrete visual tokens corresponding to a future frame (e.g., $\Delta t = 0.5s$) for a specified camera view. This pretraining forces the model to internalize spatiotemporal evolution laws, such as agent motion and ego motion from large-scale data, establishing a foundational \textit{imagination} capability without the complexity of high-level linguistic reasoning.

\noindent\textbf{SFT and RL Stages.}
For the Supervised Fine-Tuning (SFT) and Reinforcement Learning (RL) stages, we introduce a multi-step learning paradigm as illustrated in Fig.~\ref{fig: supp_data} (b) ($\approx$20k). The input is augmented with detailed vehicle kinematics (velocity, acceleration), historical trajectories, and high-level mission commands. The model output is structured into a causal reasoning sequence: it first parses the scene via <perception> and estimates a short-term <prediction>, which conditions the generation of the future <visual> frame. Crucially, the model then explicitly reasons over this imagined future in the <think> block to assess potential risks before determining the high-level <action> and regressing the precise long-term trajectory points in <answer>. This structure unifies generation and planning, allowing GRPO-based RL to optimize the consistency between the imagined future and the executed safety maneuvers.

\begin{figure*}[!t]
    \centering
    \captionsetup{type=figure}
    \includegraphics[width=1.0\linewidth]{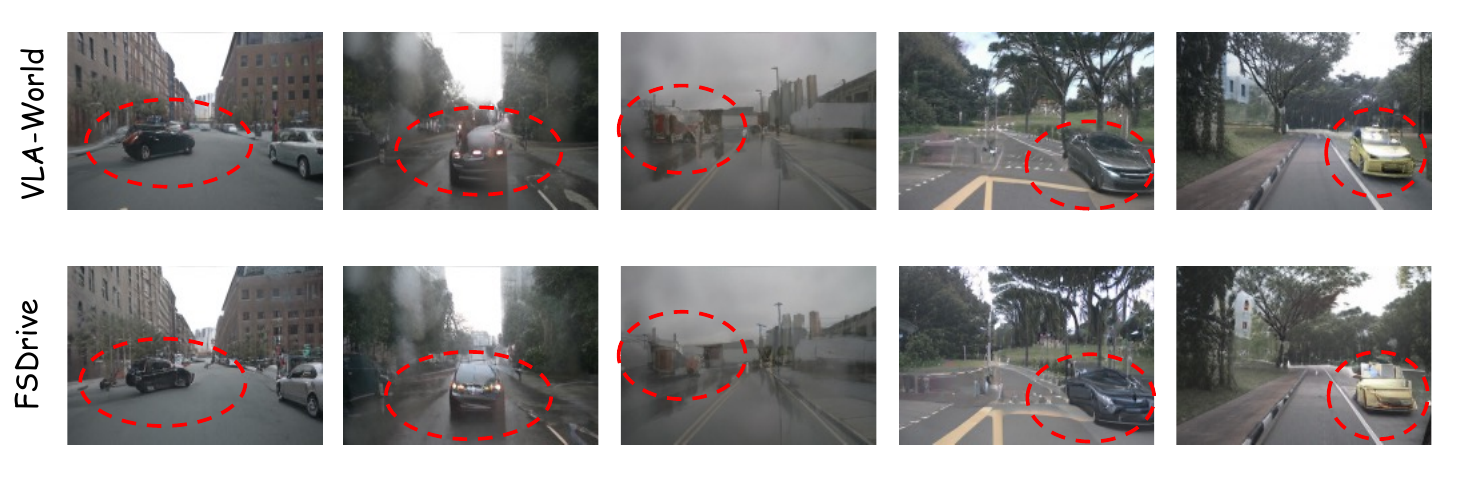}
    \caption{Comparison between our VLA-World and the state-of-the-art FSDrive~\cite{zeng2025futuresightdrive} on generating the future frame at next 0.5 seconds.}
    \label{fig: supp_gen}
\end{figure*}

\subsection{Implementation Details}
We build our model upon the Qwen-VL family~\cite{wang2024qwen2vl}. All training stages, including pretraining, supervised fine-tuning, and reinforcement learning, are conducted on 8 A100 GPUs, and inference is performed on 4 A100 GPUs. The pretraining and supervised fine-tuning stages use the LLaMA Factory framework~\cite{zheng2024llamafactory}, and the reinforcement learning stage is trained with the Easy-R1 framework~\cite{sheng2024hybridflow}. We adopt multi-view images as input and set the maximum pixel count to 524,288, with a gradient accumulation step of 2.
During pretraining, the model is trained for 30 epochs using AdamW with an initial learning rate of $5 \times 10^{-4}$ and a per-device batch size of 16. For supervised fine-tuning, we train for 12 epochs with AdamW and an initial learning rate of $1 \times 10^{-4}$. Starting from the supervised fine-tuning checkpoint, we further optimize the model for one epoch using Group Relative Policy Optimization. The policy is trained with a learning rate of $1 \times 10^{-6}$ and a global batch size of 16. To retain the behavior learned during supervised fine-tuning and ensure stable optimization, we apply a KL divergence regularization term with a coefficient of $1 \times 10^{-2}$. For each prompt, we sample 8 candidate responses to estimate the policy gradient.
A cosine learning rate scheduler with a warm-up ratio of 0.1 is applied throughout all training stages to stabilize early optimization.
We evaluate trajectory planning performance using L2 displacement error and collision rate, following widely adopted protocols in prior work~\cite{vad, vad2, hu2022stp3, wang2025omnidrive, zeng2025futuresightdrive}. UniAD~\cite{uniad} computes both metrics at each individual timestep, whereas ST P3~\cite{hu2022stp3} and VAD~\cite{vad, vad2} report the average values over all preceding timesteps. For fair comparison, we follow the respective evaluation strategies of each method. In addition, consistent with recent approaches in generative prediction~\cite{wang2023drivedreamer, wang2024driving}, we adopt the Fréchet Inception Distance to quantify the visual fidelity of synthesized future frames.

\subsection{More Discussion}

\begin{table}[]
    \centering
    \footnotesize
    \caption{Evaluation of trajectory planning L2 errors (ST-P3) on nuScenes with varying input view resolutions.}
    \setlength{\tabcolsep}{12pt}
    \renewcommand{\arraystretch}{1.0}
    \resizebox{1.0\linewidth}{!}{
    \resizebox{\linewidth}{!}{
    \begin{tabular}{lccccc}
\toprule 
\multirow{2}{*}{\textbf{Res.}} & \multicolumn{4}{c}{\textbf{L2 Error (m) $\downarrow$}} \\
\cmidrule(lr){2-5}
 & 1s & 2s & 3s & Avg. \\
\midrule
36000 & 0.03 & 0.14 & 0.98 & 0.38 \\
52884 & 0.11 & 0.27 & 0.52 & 0.30  \\
\bottomrule
\end{tabular}
    } 
}
\label{tab:supp1}
\end{table}

\begin{figure*}[!t]
    \centering
    \captionsetup{type=figure}
    \includegraphics[width=1.0\linewidth]{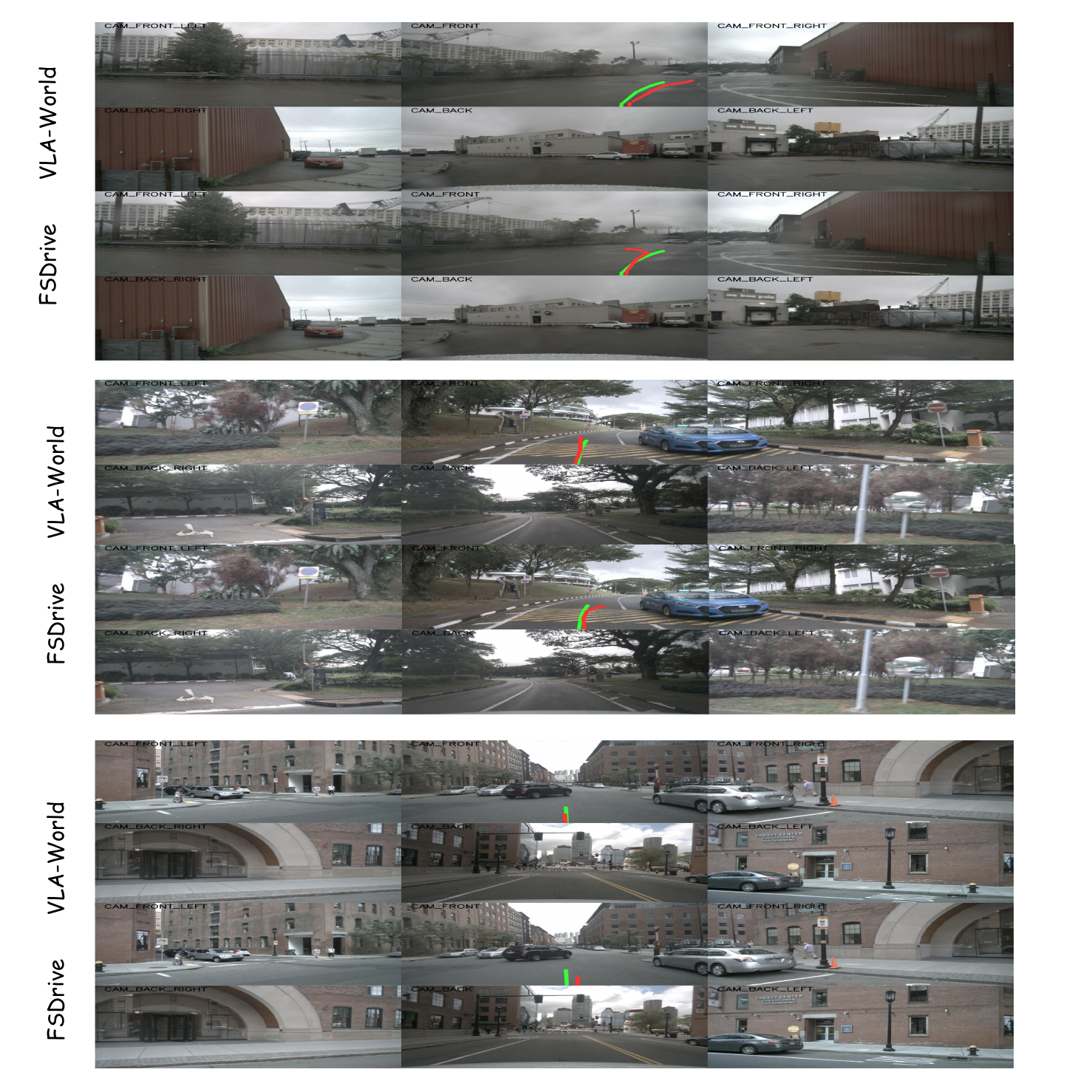}
    \caption{Comparison of 3-second future trajectory predictions generated by our VLA-World and the state-of-the-art FSDrive~\cite{zeng2025futuresightdrive}. Zoom in for a better view.}
    \label{fig: supp_plan}
\end{figure*}

\noindent\textbf{Effectiveness of Input Resolution.} 
Tab.~\ref{tab:supp1} investigates the sensitivity of our model to input view resolutions. The results demonstrate that higher resolution inputs generally yield better planning performance, particularly over longer time horizons. Although the lower resolution ($36,000$) is competitive at short intervals ($1s$), the higher resolution model ($52,884$) demonstrates superior robustness, achieving the lowest average L2 error of $0.30$m. This indicates that maintaining high-fidelity visual information is crucial for mitigating error accumulation in trajectory prediction.

\begin{table}[]
    \centering
    \footnotesize
    \caption{Evaluation of trajectory planning L2 errors (ST-P3) on nuScenes with varying model sizes.}
    \setlength{\tabcolsep}{12pt}
    \renewcommand{\arraystretch}{1.0}
    \resizebox{1.0\linewidth}{!}{
    \resizebox{\linewidth}{!}{
    \begin{tabular}{lccccc}
\toprule 
\multirow{2}{*}{\textbf{Method}} & \multicolumn{4}{c}{\textbf{L2 Error (m) $\downarrow$}} \\
\cmidrule(lr){2-5}
 & 1s & 2s & 3s & Avg. \\
\midrule
Qwen2-VL-2B & 0.11 & 0.27 & 0.52 & 0.30 \\
Qwen2.5-VL-3B & 0.05 & 0.08 & 0.76 & 0.29  \\
Qwen2-VL-7B & 0.03 & 0.03 & 0.47 & 0.18  \\
\bottomrule
\end{tabular}
    } 
}
\label{tab:supp2}
\end{table}

\noindent\textbf{Effectiveness of Model Size.} 
Tab.~\ref{tab:supp2} presents an ablation study on the effect of model size by varying the backbone among Qwen2-VL-2B, Qwen2.5-VL-3B, and Qwen2-VL-7B. The results demonstrate a clear scaling law: increasing the model capacity significantly enhances trajectory planning performance. The Qwen2-VL-7B model achieves state-of-the-art results with an average L2 error of 0.18m, outperforming the 2B and 3B variants by a substantial margin (approximately $40\%$ relative improvement). This suggests that the stronger reasoning and generalization capabilities inherent in larger parameters are essential for handling the complex causal dependencies in autonomous driving scenarios, particularly for maintaining accuracy over longer time horizons (e.g., reducing 3s error to 0.47m).

\begin{table}[]
    \centering
    \footnotesize
    \caption{Evaluation of trajectory planning L2 errors (ST-P3) on nuScenes with training strategy.}
    \setlength{\tabcolsep}{12pt}
    \renewcommand{\arraystretch}{1.0}
    \resizebox{1.0\linewidth}{!}{
    \resizebox{\linewidth}{!}{
    \begin{tabular}{lccccc}
\toprule 
\multirow{2}{*}{\textbf{Method}} & \multicolumn{4}{c}{\textbf{L2 Error (m) $\downarrow$}} \\
\cmidrule(lr){2-5}
 & 1s & 2s & 3s & Avg. \\
\midrule
w/o. Mixed & 0.27 & 0.47 & 0.73 & 0.49 \\
Qwen2-VL-2B & 0.11 & 0.27 & 0.52 & 0.30   \\
\bottomrule
\end{tabular}
    } 
}
\label{tab:supp3}
\end{table}

\noindent\textbf{Effectiveness of Training Strategy.} 
In Tab.~\ref{tab:supp3}, we conduct an ablation study to verify the contribution of our multi-task mixed dataset. We compare the full Qwen2-VL-2B model against a baseline trained without mixed data (\textit{w/o. Mixed}). The results reveal that removing the diverse supervision signals leads to a significant performance degradation, with the average L2 error increasing from 0.30m to 0.49m. This substantial gap underscores the critical role of mixed-task training (combining perception, reasoning, and planning) in learning robust feature representations, enabling the model to generalize better across varying time horizons.

\subsection{More Visualization}
We visualize the generation and trajectory planning results with the SOTA FSDrive~\cite{zeng2025futuresightdrive}. More visualization results can be found in the video demo in our supplementary materials. 

\noindent\textbf{Generation Results.}
We present a qualitative comparison of the generated 0.5s future frames in Fig.~\ref{fig: supp_gen}. As illustrated by the red-highlighted regions, the baseline method (FSDrive, bottom row) struggles to maintain object coherence during the prediction horizon. It exhibits noticeable artifacts, including geometric distortion of vehicles and a loss of high-frequency details in the background, indicating a lack of robust spatiotemporal constraints. Conversely, VLA-World (top row) demonstrates significantly improved visual fidelity. By effectively leveraging the trajectory-aware conditioning, our model preserves the structural rigidity of dynamic agents and the sharpness of the scene. The generated frames exhibit high photorealism and consistency, validating that our short-term prediction successfully mitigates the \textit{hallucination} artifacts common in pure VLA.

\noindent\textbf{Trajectory Planning.} We provide visualizations of the 3-second future trajectories predicted by VLA-World and FSDrive. As shown in Fig.~\ref{fig: supp_plan}, VLA-World produces noticeably more precise trajectory predictions than FSDrive, especially near the 3-second horizon where the deviation from the ground truth becomes minimal. This improvement stems from our paradigm: first predicting the future state, then generating the corresponding 0.5-second future frames based on that prediction, and finally reasoning over the imagined scene to refine the outcome. In contrast, FSDrive lacks such reflective and iterative reasoning capabilities, which leads to cumulative drift over longer temporal horizons.


\end{document}